\documentclass{ieeeaccess}
\usepackage{hyperref}
\usepackage{amsmath,amssymb,amsfonts}
\usepackage{algorithmic}
\usepackage{graphicx}
\usepackage{textcomp}
\usepackage{comment}
\usepackage{helvet}
\usepackage{makecell}
\usepackage{multirow}
\usepackage{booktabs}
\usepackage{soul}
\usepackage{xcolor}
\usepackage{mathptmx}
\newdimen\xfigwd
\setbox\xfigwd\hbox{\relax}

\usepackage{textcomp}
\def\BibTeX{{\rm B\kern-.05em{\sc i\kern-.025em b}\kern-.08em
    T\kern-.1667em\lower.7ex\hbox{E}\kern-.125emX}}

\begin{document}

\history{Date of publication xxxx 00, 0000, date of current version xxxx 00, 0000.}
\doi{00.0000/ACCESS.0000.DOI}

\title{A Multi-Domain Feature Fusion Framework for Generalizable Deepfake Detection Across Different Generators}

\author{
\uppercase{Amna Amjid}\authorrefmark{1},                                                
\uppercase{Sana Qadir}\authorrefmark{1}\IEEEmembership{IEEE Member},                  
\uppercase{Mehwish Fatima}\authorrefmark{1},                                            
\uppercase{and Raja Khurram Shahzad}\authorrefmark{2,3}      
}

\address[1]{School of Electrical Engineering and Computer Science (SEECS), National University of Sciences and Technology (NUST), Islamabad, Pakistan}
\address[2]{Department of Communication, Quality Management and Information Systems, Mid Sweden University, Ostersund Campus, Sweden}
\address[3]{Department of Computer Science, Electrical and Space Engineering, Lulea University of Technology, Luleå, Sweden.} 

\markboth
{A. Amjid \headeretal: A Multi-Domain Feature Fusion Framework for Generalizable Deepfake Detection Across Different Generators}
{A. Amjid \headeretal: A Multi-Domain Feature Fusion Framework for Generalizable Deepfake Detection Across Different Generators}

\corresp{Corresponding author: S. Qadir (e-mail: sana.qadir@seecs.edu.pk).}

\begin{abstract}
Deepfakes are artificially generated images, audio, or videos that threaten privacy, security, and information integrity. Detecting such content is crucial for countering disinformation, as the latest models generate highly realistic content. While spatial- or frequency-based approaches achieve good detection rates on Generative Adversarial Networks (GANs)-based generated deepfakes, they often struggle with recent diffusion model-generated images. In particular, existing approaches rarely exploit complementary multi-domain representations or systematically evaluate cross-generator robustness. To address these challenges, we propose a multi-domain deepfake detection framework called SGFF-Net (Spatial-Gradient-Frequency Fusion Network) that integrates spatial, gradient, and DWT (Discrete Wavelet Transform)-based frequency representations within a dual residual learning architecture. Experimental results show that the SGFF-Net achieves 98.95\% accuracy in intra-dataset evaluation and improves performance in both cross-model (70.46\%) and cross-paradigm (69.94\%) settings. Incorporating multi-source training and data augmentation further enhances robustness, increasing accuracy from 70.46\% to 79.80\% in cross-model evaluation, from 69\% to 78\% in cross-paradigm evaluation, and from 61.50\% to 75.80\% on real-world data. Unlike single-domain detectors, the SGFF-Net learns complementary forensic cues across spatial, gradient, and wavelet-frequency domains, resulting in greater robustness under cross-generator and cross-paradigm evaluation. The results further show that combining multi-domain representations with data diversity and augmentation substantially improves generalization, providing practical insights for developing more reliable deepfake detection systems.
\end{abstract}


\titlepgskip=-15pt

\maketitle

\section{Introduction}
\label{section:Introduction}
Recent advances in Generative Artificial Intelligence (GenAI) enable the creation of highly realistic synthetic media at unprecedented scale and accessibility. As a result, \textit{deepfakes} can now be created at scale and with minimal technical expertise, increasing their use in disinformation campaigns, impersonation attacks, fraud, blackmail, and privacy violations \cite{BBC2022DeepfakePresidents, GroupIB2025VoiceDeepfake, BBC2025AITaylorSwift}. This growing threat has elevated deepfakes from a technological concern to a broader societal and security challenge. The \textit{Global Risks Report 2025} ranks misinformation and disinformation as the fourth most significant global risk and anticipates their continued impact over the next several years \cite{WEF2025GlobalRisks}.

Detecting deepfake images, videos, and audio has become increasingly challenging with the emergence of foundation models, for example, Stable Diffusion (SD) \cite{stabilityai2025}, DALL-E \cite{ramesh2021dalle}, and StyleCLIP \cite{Patashnik2021StyleCLIP}. These models generate highly realistic synthetic content at scale, significantly expanding both the quantity and diversity of deepfakes. Beyond enabling misinformation and fraud, deepfakes also pose a growing threat to biometric authentication systems \cite{BBC2022DeepfakePresidents}.

Existing deepfake detection methods broadly fall into two categories. Cue-based approaches exploit artifacts introduced during content generation, including biological inconsistencies, spatial irregularities, temporal distortions, and frequency-domain anomalies \cite{tan2024upsampling, uhlenbrock2024palette, yu2022generalizedfeatures, liu2021spatial, jeong2022frepgan, jeong2022biHPF}. Because these artifacts are often interpretable, cue-based methods provide forensic evidence that can support detection decisions. In contrast, data-driven approaches train Machine Learning (ML) and Deep Learning (DL) models to learn discriminative representations directly from data \cite{Xie2026Survey, Pontorno2025DeepFeatureXSN, afchar2018mesonet, rossler2019faceforensics, nguyen2019capsuleforensics}.

Early detection methods primarily focused on visible spatial artifacts like abnormal textures, lighting inconsistencies, and unrealistic facial structures \cite{tan2024upsampling, uhlenbrock2024palette, yu2022generalizedfeatures, he2024rigid, liu2020texture}. Subsequent research shifted toward frequency-domain analysis, exploiting spectral patterns introduced during image synthesis \cite{qian2020frequency, li2021freqaware, liu2021spatial, jeong2022frepgan, giudice2021fighting, jeong2022biHPF}. More recently, gradient-based approaches have demonstrated the ability to capture structural inconsistencies and generator-induced traces that are less apparent in the pixel domain \cite{tan2023lgrad, zhang2025hierarchical, wu2025grenet}.

Overall, most existing methods rely on a single representation domain and rarely exploit complementary information across spatial, gradient, and frequency domains. Their performance also depends heavily on the composition of training data, augmentation strategies, and generator diversity. Consequently, detectors often struggle under distribution shift, where test images originate from unseen generators or manipulation pipelines. This limitation is particularly evident in GAN-focused detectors, which frequently rely on generator-specific visual artifacts and exhibit reduced robustness when evaluated on diffusion-generated content or real-world deepfakes \cite{Xie2026Survey, Singh2025Survey}.

Although CNN-based detectors achieve strong performance in controlled settings, their ability to learn transferable forensic representations across unseen generators remains limited \cite{Pontorno2025DeepFeatureXSN, afchar2018mesonet}. To improve robustness, several studies have incorporated frequency-domain representations, including Fast Fourier Transform (FFT) \cite{sharafudeen2025DRN, bammey2024synthbuster}, Discrete Cosine Transform (DCT) \cite{guan2023mcw}, and Discrete Wavelet Transform (DWT) \cite{abul2025fsbi, uddin2025deepfake, baru2024waveletdriven}. These transformations expose synthesis artifacts that are often less apparent in the spatial domain and have been shown to improve detection performance and cross-dataset robustness. Nevertheless, existing research focuses primarily on developing and evaluating frequency-aware detectors on GAN-generated datasets that often exhibit degraded performance when applied to diffusion-generated content, and vice versa. Despite the widespread adoption of diffusion models, few studies investigate forensic cues that transfer across both GAN and diffusion paradigms.

Recent surveys suggest that improving robustness requires moving beyond single-domain representations and leveraging complementary information from multiple feature spaces \cite{Xie2026Survey}. Motivated by this observation, we propose a multi-domain deepfake detection framework that integrates spatial, gradient, and frequency-domain representations within a Dual Residual Network (DRN) architecture. By jointly exploiting complementary forensic cues, the proposed framework, called Spatial-Gradient-Frequency Fusion Network (SGFF-Net), aims to improve detection performance and generalization across generators, paradigms, and datasets. The main contributions of this paper are summarized as follows:
\begin{itemize}
\item We develop a generalization-oriented multi-domain representation learning framework called SGFF-Net that jointly exploits spatial, gradient, and DWT-based frequency cues within a DRN to learn generator-agnostic forensic representations across diffusion and GAN paradigms. The design of SGFF-Net is guided by empirical assessment and ablation work.

\item Our proposed framework achieves strong performance across multiple evaluation settings, including an accuracy of 98.95\% for intra-dataset evaluation. It also outperforms the baseline DRN model with an accuracy of 79.80\% for cross-model evaluation and 78.12\% for cross-paradigm evaluation.

\item We perform a systematic empirical investigation to show that robust generalization depends not only on feature fusion but also on data diversity. In fact, with multi-source training and augmentation, the proposed framework achieves 75.80\% accuracy on previously unseen real-world images.
\end{itemize}

The remainder of this paper is structured as follows. Section \ref{section:RelatedWork} reviews related work on deepfake detection, including cue-based, data-driven, and diffusion-based detection methods. Section \ref{section:Methodology} describes the proposed framework. Section \ref{section:Details-of-Experiments} presents details of the experiments and implementation. Section \ref{section:Results} provides the results, while Section \ref{section:Discussion} discusses these findings. Finally, the conclusion is given in Section \ref{section:Conclusion}.

\section{Related Work}
\label{section:RelatedWork}
This section discusses different approaches for deepfake image detection, including cue-based and data-driven approaches. The cue-based detection methods attempt to exploit spatial information, gradient-domain or frequency artifacts as representations of Principal Discriminator Features (PDF) produced by the deepfake generator model. Additionally, recent deepfake detection approaches for DM-generated media are also reviewed in this section.

\subsection{Spatial-Based Deepfake-Image Detection}
\label{subsection:Spatial-Based-Deepfake-Image-Detection}
To mitigate the threats posed by sophisticated deepfakes, researchers have investigated various forgery indicators across multiple image domains, including spatial, gradient, and frequency domains. Early studies on deepfake detection used colour space \cite{tan2024upsampling, uhlenbrock2024palette, yu2022generalizedfeatures} and natural texture \cite{he2024rigid, liu2020texture} of images as PDFs. Yu et al. \cite{yu2019ganfingerprints} and Marra et al. \cite{marra2019ganfingerprint} suggested that every GAN-generated image contains hidden source-specific fingerprints in the spatial domain that can serve as reliable indicators for the digital media forensic community. Their work also explored whether these fingerprints are stable, generalizable, robust, and architecture-specific. Their results show that even very minor differences in training create different fingerprints, which indicates that fingerprints come from training dynamics, not simple artifacts. Additionally, GAN fingerprints are linked to the optimization trajectory, not merely visible artifacts. However, the studies assumed that GAN classes are known during training/testing, resulting in poor open-set generalization. 

Tan et al. \cite{tan2023lgrad} identified this limitation when they stated that there is no sufficiently generalized artifact representation for GAN-generated image detection. They hypothesised that the gradient domain of deepfake-generated images might contain discriminative regions, high-level artifacts, and generalized clues that are better suited for unseen GANs. Based on this, they proposed their LGrad (Learning on Gradients) detector, which inputs a real or fake image and passes it through a pre-trained CNN called the transformation model. This fixed transformation model is used to compute gradients that become generalized artifact representations instead of the original images. Their results proved that the gradient-based detector outperformed image-based methods and frequency-based methods. Although their method improved mean accuracy by +11.4\% against state-of-the-art (SOTA) baselines, it struggled with deepfake face-swapping methods and non-GAN generators (e.g., diffusion-based generators) because the detector mainly learnt GAN artifacts.

\subsection{Frequency-Based Deepfake-Image Detection}
\label{subsection:Frequency-Based- Deepfake-Image-Detection}
In the frequency domain, researchers investigate the impact of generative models like GANs, such as inconsistent or unusual texture in high-frequency regions or unnatural smoothness in low-frequency regions \cite{qian2020frequency, liu2021spatial, jeong2022frepgan}. These features are used as PDFs, and researchers extract them using different frequency spectral transform methods (such as FFT, DCT, and DWT) and study whether they can be used as generalized detection cues. Existing methods, e.g., the multi-feature channel domain-weighted (MCW) meta-learning framework \cite{guan2023mcw}, spatial-phase shallow learning (SPSL) \cite{liu2021spatial}, frequency-enhanced self-blended images (FSBI) \cite{abul2025fsbi}, and Frequency-aware Discriminative Feature Learning framework (FDFL) \cite{li2021freqaware}, exploit frequency extraction methods to capture abnormalities in the frequency domain and use them as detection cues. Uddin et al. proposed a deepfake face detection approach that combines the output of the multi-level DWT and vision transformer to obtain frequency features, distinguishable information, and global forgery cues \cite{uddin2025deepfake}. Giudice et al. analytically demonstrated the nature of multimedia content as real or deepfake using DCT to detect so-called GAN-specific frequencies (GSFs) \cite{giudice2021fighting}. 
Their method models the 63 Alternating Current (AC) coefficients, corresponding to the frequency-detail components of 8×8 DCT blocks, with a Laplacian distribution and extracts the $\beta$ scale parameter, which measures the spread or variation of each frequency component. The results demonstrate that the $\beta$ statistics inferred from the distribution of AC coefficients are key to recognizing the data generated by the GAN engine and serve as explainable fingerprints of GAN-generated images.


More recently, Sharafudeen \& Chandra proposed a framework that consists of a Dual Residual Network (DRN) that effectively captures local and global features using deep residual connections \cite{sharafudeen2025DRN}. This framework focuses on high-frequency analysis using  FFT to detect subtle inconsistencies in synthetic images. To detect both subtle and advanced forgery traces, Qian et. al. proposed a Frequency in Face Forgery Network ($F^3$-Net) that utilizes a dual-path architecture designed to reveal manipulation footprints through the collaborative analysis of decomposed frequency features and local probabilities \cite{qian2020frequency}. Jeong et al. proposed a robust deepfake detection system called BiHPF that consists of two high-pass filters (HPFs) \cite{jeong2022biHPF}. One filter is a frequency-level HPF that amplifies the magnitude, and the second is a pixel-level HPF to emphasize the surrounding background pixel values. These two HPFs jointly increase the visibility of discriminator artifacts. In another similar work, Jeong et al. observed that the generalization capability of detection models weakens because the frequency artifacts in generated images are very sensitive to the distribution of the generator and object categories \cite{jeong2022frepgan}. They used this insight to develop FrePGAN - a system that improves model generalization by introducing frequency-level perturbation maps to remove the effect of the frequency-level artifacts.

\subsection{Data-driven Deepfake-Image Detection}
\label{subsection:Data-driven Deepfake Detection}
Although the studies that only focused on the frequency domain show satisfactory performance for intra-dataset evaluation carried out using GAN-based datasets, there is a drastic performance degradation when the evaluation is carried out in cross-dataset settings (especially for the recent diffusion models), which indicates that this approach has poor generalizability. Another issue with GAN-generated faces is that it is possible to remove the GAN fingerprints from the frequency domain through adversarial techniques, thus disabling detectors based on such cues \cite{Xie2026Survey, dasgupta2025attentionenhanced}. To overcome such limitations, research has shifted towards data-driven deepfake detection instead of manually extracting handcrafted cues from spatial or frequency domains. Data-driven deepfake detection utilizes Deep Neural Networks (DNNs) and massive quantities of training data instead of relying on specific feature extraction methods to extract PDFs. For instance, Marra et al. conducted a comparative study of deepfake detection using different DNNs (e.g., InceptionNet, DenseNet, and XceptionNet) trained using a large dataset \cite{marra2018socialgan, zhu2017cyclegan}. Even though their data analysis showed that XceptionNet outperformed all the other models, they did not explore generalization of the DNNs.

Ameer et al. investigated whether data augmentation genuinely improves the deepfake detection rate and generalizability of data-driven models \cite{ameer2025augmentation}. They compared three different CNN models (XceptionNet, ResNet-50, and EfficientNet-B3) and two transformer models (ViT-B/16 and ConvNeXt-Small), both with and without augmentation, using the GAN-based Deepfake Detection Challenge (DFDC) dataset \cite{dang2020detection}. Their cross-model experimental results for XceptionNet (89\%) and ViT-B/16 (75\%) proved that data augmentation improves cross-model generalization more in CNNs compared to transformer models.

Recently, Pontorno et al. used data-driven self-supervised contrastive learning to propose a DeepFeatureX-SN framework \cite{Pontorno2025DeepFeatureXSN}. Architecturally, it leveraged tripartite Siamese DenseNet161 extractors, working in parallel to counter the difficulties of feature extraction. Each extractor was tasked with deriving convolutional features from only one type of image: real, GAN, or DM-generated. The evaluation of the framework showed very good performance by achieving 97.70\% accuracy for binary classification (for the same dataset evaluation). During generalization assessment, the framework achieved an accuracy of 74.45\% for unseen GAN-generated images and 57.24\% on unseen DM-generated images. While DeepFeatureX-SN performance was superior compared to existing state-of-the-art methods, it had high computational requirements and relatively low performance for unseen DM-generated images. 

\begin{table*}[!t]
\caption{Comparison of representative deepfake and synthetic image detection methods in terms of feature domain, methodology, key findings, and limitations.}
\label{tab:related_work}
\centering
\footnotesize
\renewcommand{\arraystretch}{1.2}
\begin{tabular}{l l l l l}
\toprule
\textbf{Ref.} & \textbf{Feature / Domain} & \textbf{Methodology} & \textbf{Key Findings} & \textbf{Limitations} \\ \midrule
\cite{tan2023lgrad} & Spatial / Gradient & \makecell[l]{LGrad using pre-trained CNN,\\gradient representation,\\and classifier} & \makecell[l]{Gradient artifacts improved\\cross-model generalization\\by up to 11.4\%} & \makecell[l]{Weak on non-GAN deepfakes;\\sensitive to JPEG compression\\and noise} \\ \midrule
\addlinespace
\cite{dasgupta2025attentionenhanced} & Data-driven & \makecell[l]{CNN + Multi-Head\\Self-Attention (MHSA)\\with attention heatmaps} & \makecell[l]{Achieved 94--98\% accuracy\\and up to 99.75\% AUC\\using local and global features} & \makecell[l]{Poor generalization on\\unseen datasets\\and generators} \\ \midrule
\addlinespace
\cite{Pontorno2025DeepFeatureXSN} & Data-driven & \makecell[l]{Siamese Network with\\contrastive learning and\\CNN classifier (DeepFeatureX-SN)} & \makecell[l]{Achieved 97.29\% accuracy\\and improved cross-generator\\performance (67.40\%)} & \makecell[l]{Limited robustness on\\unseen generators;\\high computational cost} \\ \midrule
\addlinespace
\cite{sharafudeen2025DRN} & Frequency & \makecell[l]{FFT-based frequency forensics\\with Dual Residual Network (DRN)} & \makecell[l]{Residual learning captured\\frequency fingerprints and\\improved detection performance} & \makecell[l]{Strong dataset bias and\\limited leave-one-generator-out\\generalization} \\ \midrule
\addlinespace
\cite{abul2025fsbi} & Frequency & \makecell[l]{Self-Blended Images (SBI)\\augmentation with\\DWT-based features} & \makecell[l]{DWT exposed subtle forgery\\artifacts and improved\\cross-dataset performance} & \makecell[l]{Limited cross-generator\\and diffusion-model\\evaluation} \\ \midrule
\addlinespace
\cite{corvi2023DM} & Frequency & \makecell[l]{CNN-based frequency analysis\\for GAN and diffusion\\image detection} & \makecell[l]{Frequency artifacts provided\\strong discriminative\\forensic cues} & \makecell[l]{Poor generalization to\\unseen generators\\and datasets} \\ \midrule
\addlinespace
\cite{sha2023defake} & Spatial + Semantic & \makecell[l]{Multimodal diffusion attribution\\using image and text-prompt\\embeddings} & \makecell[l]{Captured diffusion-specific\\fingerprints using\\semantic cues} & \makecell[l]{Closed-set limitation;\\requires prompts and lacks\\real-world robustness} \\ \midrule
\addlinespace
\cite{bammey2024synthbuster} & Frequency & \makecell[l]{FFT high-pass filtering with\\spectral peak extraction\\and gradient boosting} & \makecell[l]{Effective on multiple\\diffusion models with\\partial compression robustness} & \makecell[l]{Sensitive to domain shift\\and unseen diffusion models} \\
\bottomrule
\end{tabular}
\end{table*}

\subsection{Diffusion Model Deepfake-Image Detection}
\label{subsection:Diffusion-Model-Deepfake-Detection}
The last few years have seen the quality and realism of deepfake technology evolve with a shift from GANs to DMs. This can be seen in the adoption of mainstream image synthesis models, e.g., Denoising Diffusion Implicit Models (DDIM) \cite{song2020ddim}, Denoising Diffusion Probabilistic Models (DDPM) \cite{nichol2021improvedddpm}, SD \cite{stabilityai2025}, Midjourney, DALL·E 3 \cite{ramesh2021dalle}.
These models allow users to achieve real-time text-to-image generation with only prompts, further lowering the barrier to using deepfakes. However, most existing deepfake detection methods focus largely on distinguishing real content from GAN-generated content. In this section, the few notable studies that specifically tackle the detection of deepfake images generated by DMs are discussed.

In 2023, Corvi et al. investigated the problem by attempting to understand how difficult it is to distinguish synthetic images generated by diffusion models from real ones and whether current SOTA detectors were suitable for the task \cite{corvi2023DM}. They retrained the existing architecture of Gragnaniello et al. \cite{gragnaniello2021gan} on DM-generated images and used the findings to highlight that synthetic images generated by GANs and DMs have specific fingerprints that depend on the architecture and the parameters of the generative network used to generate them.

In the same year, Sha et al. proposed DE-FAKE, a method based on an ML classifier for diffusion model detection on four popular text-to-image architectures (including DALLE 2 and Latent Diffusion) 
\cite{sha2023defake}. Their empirical results showed that deepfake images generated by different diffusion models can be distinguished from real ones because of the existence of common artifacts in diffusion-generated images using frequency-domain analysis. Additionally, they showed that different diffusion models leave unique fingerprints in their generated images, due to which deepfake images can be effectively attributed to their source models.

Bammey investigated synthetic image detection and proposed a method based on spectral analysis for diffusion models \cite{bammey2024synthbuster}. He extracted the artifacts of residual images left during the Fourier transformation in the diffusion process and used some manually selected frequency components for the detection process.
A key contribution of the work is the demonstration that a simple cross-difference high-pass filter can reveal diffusion-related frequency artifacts more effectively than prior residual extraction approaches, enabling their detection at the individual-image level.

Chen et al. developed a masked conditional DM for deepfake detection and specifically studied the use of augmentation for improving generalization \cite{chen2024mcdm}. Instead of directly swapping faces and using traditional augmentations (flip, rotation, blur, etc.), they randomly masked facial regions and used a conditional diffusion model to reconstruct or manipulate the masked areas, generating a diverse range of forged faces with subtle artifacts. During experimentation, the data showed clear improvement in cross-dataset generalization. For example, the AUC for the Celeb-DF dataset improved from 73.12\% (without augmentation) to 78.02\% using their diffusion-based augmentation.

\vspace{0.25cm}

Table~\ref{tab:related_work} presents a summary of notable image deepfake detection methods discussed above in terms of feature representation, methodology, key findings, and limitations. It can be noted that spatial and frequency artifacts provide strong discriminative cues for detection of AI-generated images \cite{tan2023lgrad, corvi2023DM}. However, most existing approaches rely heavily on either spectral artifacts or spatial-domain inconsistencies. This reliance on representations from a single domain leads to limited generalization capability, especially for unseen generative models, by not integrating complementary information from the other domains. A few hybrid approaches have been developed, such as DE-FAKE, which improve detection by incorporating prompt semantics, but they depend on auxiliary textual information that might not always be available in real-world scenarios \cite{sha2023defake}. Additionally, most existing frequency analyses are limited, meaning that they primarily focus on analysing the high Fourier spectrum artifacts and frequency peaks \cite{sharafudeen2025DRN, bammey2024synthbuster}. In other words, they lack localized multi-resolution frequency learning. Also, recent studies have emphasized the importance of robust feature representations and augmentation strategies for improving generalization, but these techniques have not been sufficiently explored \cite{khan2024generalization, Xie2026Survey}. Furthermore, many existing methods demonstrate strong intra-dataset performance but suffer considerable performance degradation during cross-dataset and cross-generator evaluation \cite{sharafudeen2025DRN, tan2023lgrad}. This paper aims to address the above-mentioned research gaps by achieving the following two research objectives:
\begin{enumerate}
\item Design and develop a deepfake detection framework called SGFF-Net that integrates spatial (RGB), gradient, and DWT-based frequency representations within a Dual Residual Learning (DRN) architecture for improved feature representation and generalization.
\item Evaluate the generalization of SGFF-Net in cross-model, cross-paradigm, multi-source, and real-world scenarios (including with and without augmentation and diverse training strategies).
\end{enumerate}

\begin{figure*}[t]
    \centering
    \includegraphics[width=0.75\textwidth]{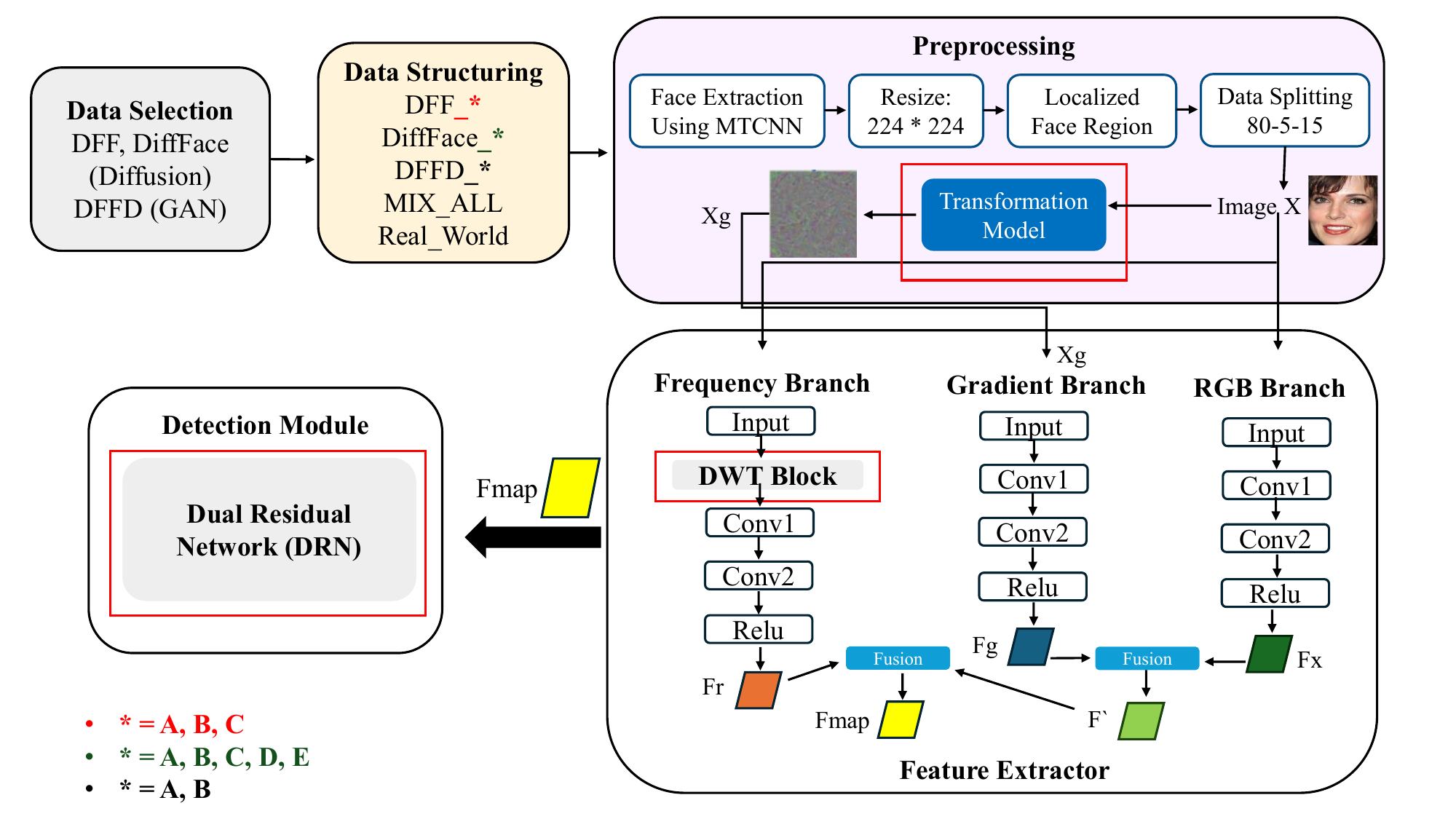} 
    \caption{Overview of Spatial-Gradient-Frequency Fusion Network (SGFF-Net): the proposed multi-domain deepfake detection framework. RGB, gradient, and DWT-based frequency representations are extracted in parallel and fused before classification using the Dual Residual Network (DRN).}
    \label{fig:architecture}
\end{figure*}

\section{Spatial-Gradient-Frequency Fusion Network for Generalizable Deepfake Detection}
\label{section:Methodology}

Figure~\ref{fig:architecture} shows the architecture of our proposed framework, Spatial-Gradient-Frequency Fusion Network (SGFF-Net) for Generalizable Deepfake Detection. It consists of four stages: data selection and structuring, preprocessing, feature extraction, and DRN-based detection. We start by selecting three distinct publicly available datasets: DeepFakeFace (DFF)\footnote{\url{https://github.com/OpenRL-Lab/DeepFakeFace}}, Diffusion Face (DiffFace)\footnote{\url{https://github.com/Rapisurazurite/DiffFace}}, and Diverse Fake Face Dataset (DFFD) \footnote{\url{https://cvlab.cse.msu.edu/dffd-dataset.html}} (GAN-based), and use the images/instances of these datasets to create multiple structured datasets for structured evaluation. The preprocessing model takes all images (including the generated ones) and does face extraction and localization using Multi-task Cascaded Convolutional Networks (MTCNN). Furthermore, with each original image $X$, we compute a gradient map $X_g$  by a transformation model (based on ResNet-50). All original images and their corresponding maps are passed into the feature extractor. The feature extractor consists of three separate components: RGB (Spatial), gradient, and frequency, each using two sets of convolution for generating feature maps $F_x$, $F_g$, and $F_r$ respectively. We then combine them to generate a final feature map $F_{map}$ of size (192, 224, 224). The final feature map $F_{map}$ serves as an input for the detection. We assess the performance of our framework empirically with multiple systematic settings with an aim at generalization.


\subsection{Dataset Selection and Data Structuring}
\label{subsubsection:Datasets}

Table~\ref{tab:datasets} shows a summary of three datasets along with their fake generator methods. DFF and DiffFace are diffusion-based datasets, while DFFD is a GAN-based dataset. We divide these datasets into multiple subsets to support different evaluation scenarios: (1) Intra-dataset, (2) Cross-model, (3) Cross-paradigm, (4) Multi-source, and (5) Real-world evaluation. The DFF dataset is divided into three subsets (DFF\_A, DFF\_B, and DFF\_C) that are mainly used for diffusion-based intra-dataset evaluation. The DiffFace dataset contains synthetic images generated using multiple diffusion models, including Perception Prioritized Training (P2), DDPM \cite{ho2020ddpm}, Pseudo Numerical Methods for Diffusion Models (PNDM) \cite{liu2022pndm}, Latent Diffusion Models (LDM), DDIM \cite{song2020ddim}, and Ablated Diffusion Models (ADM) \cite{rombach2022ldm}. For this reason, DiffFace is divided into multiple subsets (A, B, C, D, and E), so that each subset can be used for cross-model evaluation where the model is trained on the DFF\_C dataset. The DFFD dataset is further divided into DFFD\_A and DFFD\_B. The DFFD\_A dataset includes fake images generated using Progressive Growing of GANs versions 1 and 2 (PGGAN V1, PGGAN V2) and FaceApp. The DFFD\_B dataset includes fake images from StyleGAN and StarGAN. The former is used for cross-paradigm testing (with training done using diffusion models), while the latter is used for cross-model testing (with training done on GAN-based datasets). A combined dataset is created from the mixture of DFF\_C, DiffFace\_A, and DFFD for multi-source evaluation. Finally, to assess SGFF-Net's robustness in a real-world context, 400 unseen images are randomly sourced from the Internet to create the Real\_World dataset. The details of these constructed evaluation datasets are presented in Table~\ref{tab:evaluation_datasets}. To compile the Real\_World dataset, 200 authentic (real) images and 200 AI-generated images were sourced from the internet. The search subjects included a wide variety of prominent celebrities, public figures, and media personalities. For each selected individual, real facial images were pulled from publicly accessible web sources and image repositories. To obtain AI-generated samples, the same celebrity names were used as search queries combined with keywords such as ``I-generated'', ``AI portrait'', ``synthetic image'', ``deepfake'', and related terms.  

\begin{table*}[!th]
\caption{Characteristics of the original datasets used in this study, including generation methods, source models, and real-image distributions.}
\label{tab:datasets}
\centering
\renewcommand{\arraystretch}{1.2}
\begin{tabular}{l l l l l l}
\toprule
\textbf{Dataset} & \textbf{Generation Method} & \textbf{Model} & \textbf{No. of Fake Images} & \textbf{Source of Real Images} & \textbf{No. of Real Images} \\ \midrule

DFF & Diffusion & SD v1.5 text-to-image & 30,000 & IMDb-Wiki Dataset & 30,000 \\
DFF & Diffusion & SD Inpainting & 30,000 & IMDb-Wiki Dataset & 30,000 \\
DFF & Diffusion & InsightFace & 30,000 & IMDb-Wiki Dataset & 30,000 \\ \midrule

DiffFace & Diffusion & \makecell[l]{Unconditional\\(P2, DDPM, DDIM,\\PNDM, LDM)} & 150,000 & CelebA & 30,000 \\
DiffFace & Diffusion & SD v2.1 text-to-image & 90,000 & CelebA & 30,000 \\
DiffFace & Diffusion & SD v2.1 image-to-image & 90,000 & CelebA & 30,000 \\
DiffFace & Diffusion & SD v1.5 text-to-image & 90,000 & CelebA & 30,000 \\
DiffFace & Diffusion & SD v1.5 image-to-image & 90,000 & CelebA & 30,000 \\
DiffFace & Diffusion & DiffSwap & 30,000 & CelebA & 30,000 \\
DiffFace & Diffusion & Inpainting & 30,000 & CelebA & 30,000 \\
DiffFace & Diffusion & Internet & 30,000 & CelebA & 30,000 \\ \midrule

DFFD & GAN & PGGAN V1 \& V2 & 200,000 & \makecell[l]{CelebA +\\FFHQ} & 781,727 \\
DFFD & GAN & StarGAN & 80,000 & CelebA & 2,000 \\
DFFD & GAN & FaceApp & 12,000 & FFHQ & 4,000 \\
DFFD & GAN & StyleGAN & 100,000 & \makecell[l]{Random latent\\generation} & -- \\

\bottomrule
\end{tabular}

\end{table*}

\begin{table*}[t]
\caption{Structured evaluation datasets derived from DFF, DiffFace, and DFFD to support intra-dataset, cross-model, cross-paradigm, multi-source, and real-world evaluation.}
\label{tab:evaluation_datasets}
\centering
\renewcommand{\arraystretch}{1.2}
\begin{tabular}{l l l l l}
\toprule
\textbf{Type of Evaluation} & \textbf{Dataset Created} & \textbf{Originating Dataset} & \makecell[c]{\textbf{No. of Real Images}\\\textbf{(Source Dataset)}} & \makecell[c]{\textbf{No. of Fake Images}\\\textbf{(Source Dataset / Model)}} \\ \midrule

Intra-dataset & DFF\_A & Deep Fake Faces (DFF) & 23,137 (IMDb-Wiki) & \makecell[l]{23,137\\(SD v1.5 text-to-image)} \\
Intra-dataset & DFF\_B & Deep Fake Faces (DFF) & 23,137 (IMDb-Wiki) & 23,137 (SD Inpainting) \\
Intra-dataset & DFF\_C & \makecell[l]{Mix of DFF\_A\\and DFF\_B} & 46,274 (IMDb-Wiki) & \makecell[l]{46,274\\(SD v1.5 text-to-image\\+ SD Inpainting)} \\ \midrule

Cross-model & DiffFace\_A & Diffusion Face (DiffFace) & 11,019 (CelebA) & \makecell[l]{11,019\\(DDPM, DDIM,\\PNDM, and LDM)} \\
Cross-model & DiffFace\_B & Diffusion Face (DiffFace) & 13,627 (CelebA) & 13,627 (ADM) \\
Cross-model & DiffFace\_C & Diffusion Face (DiffFace) & 13,627 (CelebA) & 13,627 (DDPM) \\
Cross-model & DiffFace\_D & Diffusion Face (DiffFace) & 13,627 (CelebA) & 13,627 (LDM) \\
Cross-model & DiffFace\_E & Diffusion Face (DiffFace) & 13,627 (CelebA) & 13,627 (PNDM) \\
Cross-model & DFFD\_B & \makecell[l]{Diverse Fake Face\\Dataset (DFFD)} & 10,413 (CelebA) & \makecell[l]{10,413\\(StyleGAN and StarGAN)} \\ \midrule

Cross-paradigm & DFFD\_A & \makecell[l]{Diverse Fake Face\\Dataset (DFFD)} & 23,137 (CelebA) & \makecell[l]{23,137\\(PGGAN v1/v2\\and FaceApp)} \\ \midrule

Multi-source & MIX\_ALL & \makecell[l]{Mix of DFF\_C,\\DiffFace\_A, and DFFD} & 90,843 & 90,843 \\

Real-world & Real\_World & Internet & 200 & 200 \\

\bottomrule
\end{tabular}

\end{table*}


\subsection{Preprocessing}
The Preprocessing module performs face region extraction and localization using MTCNN and resizes images to $224 \times 224$ pixels. The resulting images are stored in RGB JPEG (Joint Photographic Experts Group) format. The dataset is then split, with 80\% of the dataset used for training, 5\% for validation, and the remaining 15\% for testing. Another key part of the pre-processing module is that the original RGB image, $X$, is also processed by a transformation model to produce a gradient map $X_g$. The detailed description and the mathematical function for creating gradient maps are discussed below.


Prior studies have shown that CNN gradient responses capture generator-induced artifacts that transfer more effectively across synthetic image generation methods than raw pixel representations \cite{tan2023lgrad,zhang2025hierarchical,wu2025grenet}. Following this observation, a ResNet-50-based transformation model is trained and subsequently used to generate gradient maps, where pixel magnitudes reflect their contribution to the real-versus-fake prediction.
Mathematically, a dataset is defined $X = \{(x_i, y_i)\}_{i=1}^{N}$, where $x_i \in \mathbb{R}^{3 \times H \times W}$ represents an RGB image, $y_i \in \{0,1\}$ denotes the class label (\texttt{Real} or \texttt{Fake}), $N$ is the total number of samples, $H$ represents the height of the image, and $W$ denotes the width of the image. Also, given that the trained transformation model is denoted as $G(\cdot)$, each input image, $x_i$, is passed through the trained transformation model to obtain the corresponding output representation: \begin{equation}z_i = G(x_i) \end{equation} where $z_i$ is the output vector of the transformation model G($.$). Then, the sum of gradients ($z_i$) is calculated with respect to the input  $x_i$. Finally, the gradient of the summed output with respect to the input image is computed as:
\begin{equation}
M_i = \frac{\partial \sum z_i}{\partial x_i}
\end{equation}

\subsection{Feature Extractor}
\label{subsection:Feature Extractor}
The output of the transformation model ($X_g$) and the original RGB image ($X$) are directly input into the feature extractor module. This module consists of three separate branches: RGB (Spatial), gradient, and frequency. Two sets of convolution operations are performed separately on each branch to obtain feature maps $F_x$, $F_g$, and $F_r$. The details of the feature representation method and the internal working of the DWT module are discussed in the rest of this section. 


\subsubsection{Selection of Frequency Representation}
\label{subsubsection:Selection-of-Frequency-Representation-Method}
\begin{figure*}[t]
  \centering
  \includegraphics[width=0.75\textwidth]{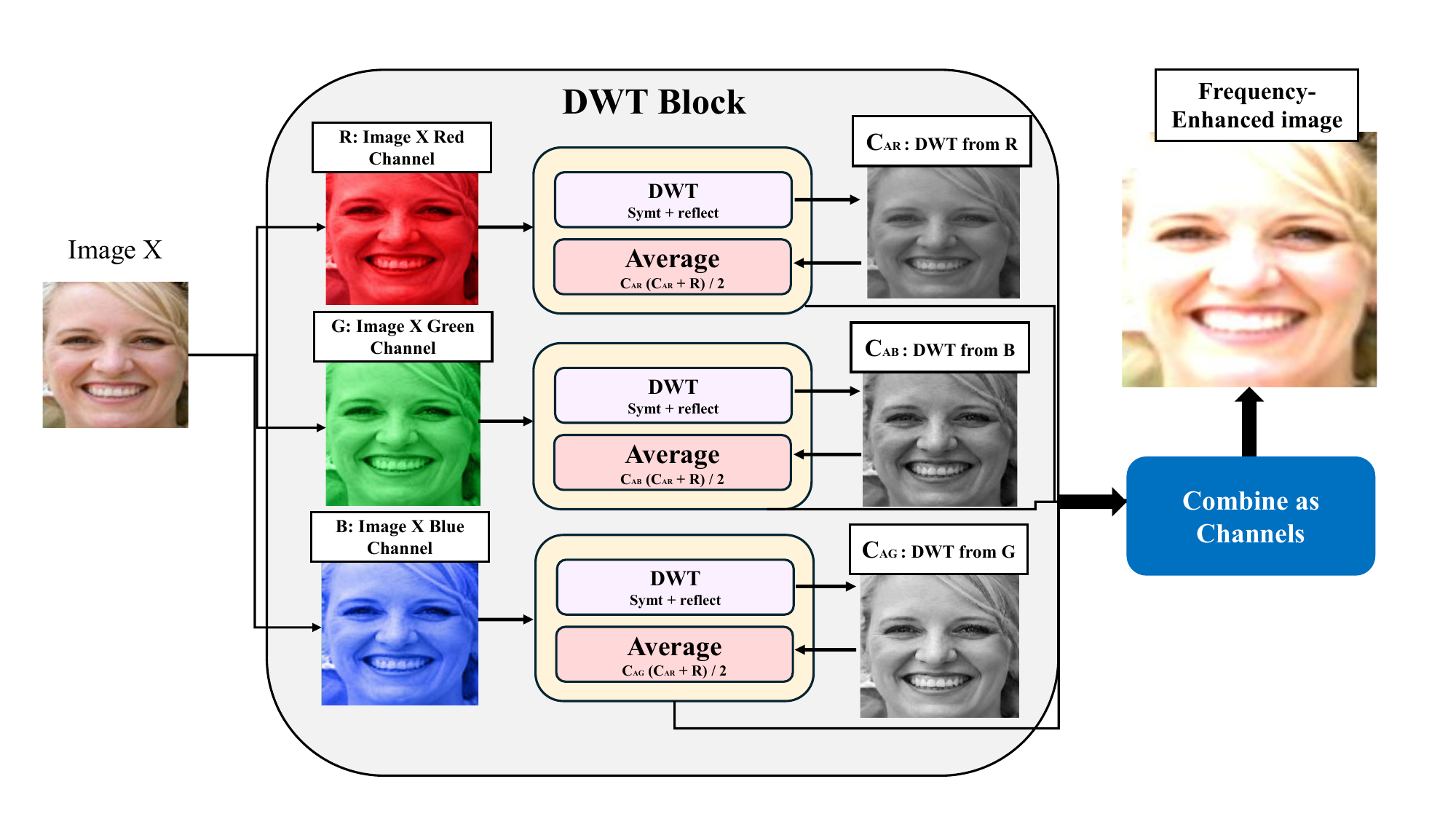}
  \caption{DWT-based frequency extraction module. Each RGB channel is decomposed using DWT, reconstructed into a frequency-enhanced representation, and forwarded to the frequency feature extractor branch.}
  \label{fig:DWTarchitecture}
\end{figure*}

The choice of frequency representation method for frequency-enhanced images within the DWT block is made based on an empirical evaluation of different frequency representations (i.e., FFT, DCT, DWT) to determine which had the best performance for detecting DM-generated deepfake images. Two categories of datasets are created: one is the original RGB, and the second is its grayscale version. Then, for each version, the image is converted to the frequency domain using 2D FFT, DCT, and DWT A high-pass circular mask is applied and reconstructed in the case of FFT and DCT. For DWT, the HH (High-High) components are extracted and resized to the original resolutions. The selection of high-frequency components is motivated by the results of existing literature \cite{sharafudeen2025DRN, abul2025fsbi, bammey2024synthbuster, guan2023mcw}, which demonstrate that synthesis and reconstruction artifacts are more prominent in high-frequency regions. To determine the most effective representation, comparative experiments are carried out, and the results consistently demonstrated that RGB DWT-based representations achieved superior performance across most evaluation metrics, particularly in detecting diffusion-based manipulations (see Table \ref{tab:frequency_ablation}). Based on these findings, DWT is selected as the primary frequency transformation method in the  DWT-block. The architecture of this module is illustrated in Fig. \ref{fig:DWTarchitecture}.

\subsubsection{DWT Transformation}
\label{subsubsection:DWT-Transformation-Module}
Initially, each input image is represented in RGB format. Instead of feeding this image directly into a convolutional network, the frequency branch first decomposes it into multiple frequency bands. A single-level two-dimensional DWT is applied to each channel. The DWT module decomposes the input signal into four distinct subbands: LL (low-frequency in both directions), LH (low-frequency horizontal, high-frequency vertical), HL(high-frequency horizontal, low-frequency vertical), and HH (high-frequency in both directions). This is performed channel-wise, meaning each channel (the red, green, and blue) is processed independently to preserve channel-specific frequency characteristics. The LL subbands represent a low-frequency approximation of the image and capture its global structure, such as smooth regions and overall shape. The LH, HL, and HH subbands capture high-frequency details corresponding to horizontal edges, vertical edges, and fine-grained textures or noise, respectively. To evaluate the contribution of each component, multiple configurations are tested, including individual subbands (LL, HH), partial combinations (LL+HH), and full fusion of all subbands (LL+LH+HL+HH). Based on the comparative results (see Figs.~\ref{fig:f1_subband} and ~\ref{fig:acc_subband}), the fusion of all subbands (LL+LH+HL+HH configuration) was selected, as it had the highest accuracy and F1 score. 

Additionally, the choice between different DWT wavelet families depends on the specific characteristics of the deepfake images and the requirements of the detection task. Therefore, the Haar, Symlet, and Biorthogonal wavelet types are assessed for deepfake detection. The best results are obtained using the Symlet wavelet type (see Table~\ref{tab:wavelet_ablation}). When applying DWT, borders need handling; for this purpose, different signal extension modes are used. Most commons modes are zero padding, symmetric, and anti-reflect. All of these modes are tested (see Table~\ref{tab:wavelet_ablation}), and because the highest score is achieved using the reflective mode, this mode is implemented in the final design. 

The resulting combined subband is reconstructed channel-wise, and the processed channels are merged to produce a three-channel frequency-enhanced image (see Fig. \ref{fig:DWTarchitecture}). This frequency-transformed image is subsequently converted into a tensor representation and passed through a dedicated convolutional branch to give $F_r \rightarrow (64, 224, 224)$. Finally, using channel-wise fusion, $F`$ and $F_r$ are combined to form a final feature map $F_{map}$ of size $(192, 224, 224)$, which is then fed into the detection module.


\subsection{Detection Network}
\label{subsection:Detection-Module-Design}
\begin{figure*}[t]
  \centering
  \includegraphics[width=0.75\textwidth]{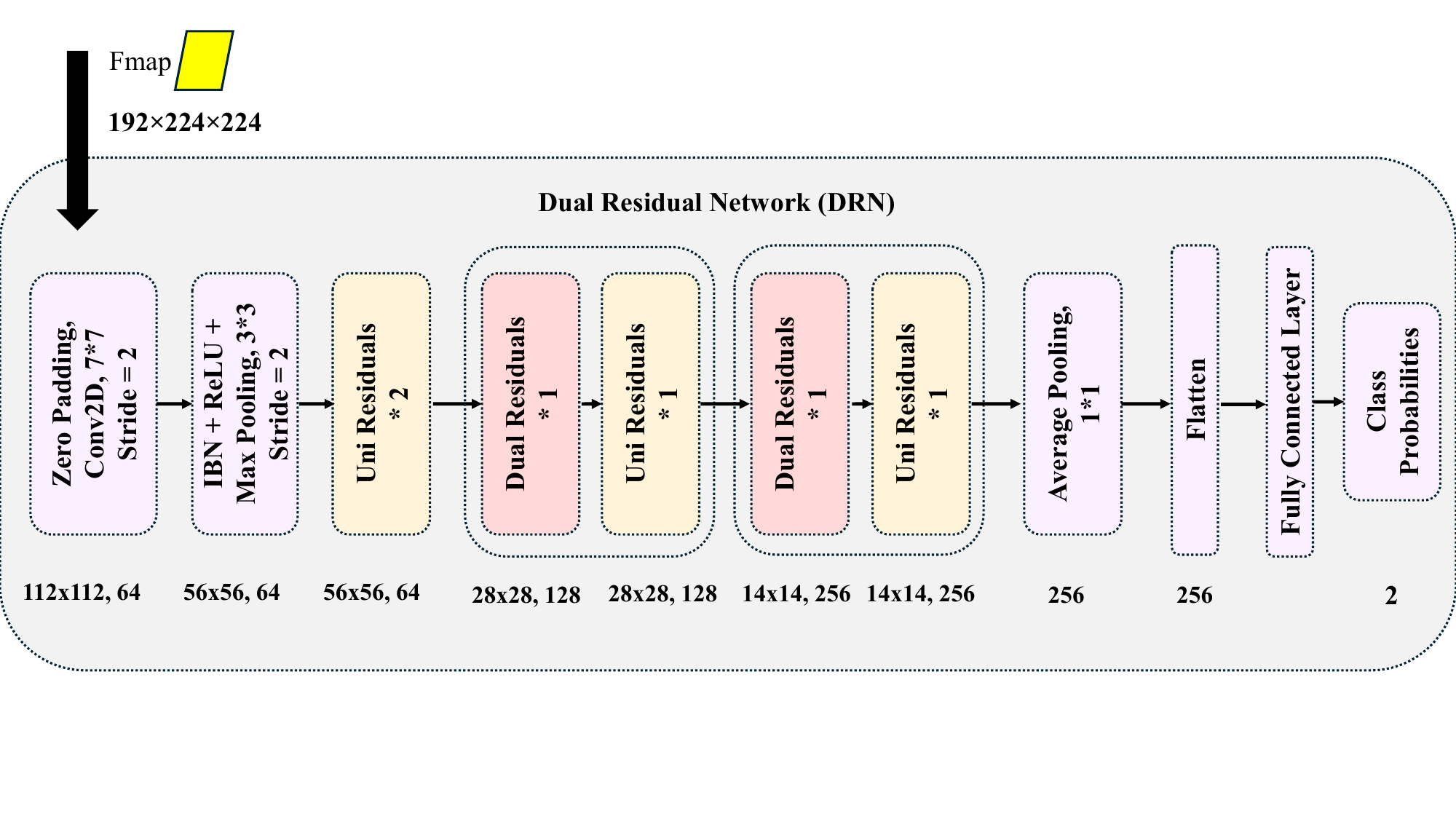}
  \caption{Architecture of the Dual Residual Network (DRN) used for classification. The network combines local and global residual paths with IBN normalization to learn robust forgery representations from fused multi-domain features.}
  \label{fig:DRN}  
\end{figure*}

Inspired by the foundation models of ResNet \cite{he2016resnet},  and the model developed in \cite{sharafudeen2025DRN}, a Dual Residual Network (DRN) structure is selected for the architecture of the detection module. Specifically, it is designed to efficiently learn discriminative representations, combining residual learning, dual-path feature learning, and Instance-Batch Normalization (IBN). It has few parameters and operates on fused multi-domain features, where the local path facilitates inter-channel feature interaction across colour, gradient, and frequency domains, while the global path captures spatial correlations of forgery artifacts. Additionally, it is lightweight because it incorporates IBN in early layers, reducing network depth and simplifying the classification head. These modifications significantly reduce computational complexity while improving generalization across diffusion and GAN-based deepfake datasets. Figure \ref{fig:DRN} visually represents the design of the detector module.

The DRN begins with a convolutional layer that transforms the initial 192-feature map into 64 channels using a $7 \times 7$ kernel with a stride of 2. This operation reduces the spatial information from the fused features while capturing the global contextual information. The output from this layer is then passed through an IBN layer followed by a ReLU activation function. The IBN layer divides the feature channel into two groups; one half is normalized using Instance Normalization, and the other half using Batch Normalization. Following the initial encoding, the network applies a sequence of uni-residual blocks. Each block consists of a convolutional layer followed by normalization and ReLU activation. For robustness against face variation, the IBN is used in the initial layer of the uni-residual block.

Dual-residual paths are also added to the model to capture fine-grained and contextual information. The initial path is a local path that captures fine-grained details and artifact patterns using $1 \times 1$ convolutions, and the global path captures broader contextual information using stacked $3 \times 3$ convolutions. The outputs of these two paths are combined along with a residual connection from the input, and the result is passed through a ReLU activation. Global feature aggregation is done by passing the resulting feature map through a global average pooling layer to produce a compact feature vector of size 256. Finally, the pooled feature vector is fed into a fully connected layer, which maps the 256-dimensional feature vector to a 2-dimensional output corresponding to the binary class. A softmax function is applied to the output logits for final prediction, followed by the selection of the class with the highest probability.

\subsection{Augmentation Techniques}
\label{subsection:Augmentation-Techniques}
Recent comparative analyses demonstrate that augmentation strategies can enhance cross-dataset generalization performance in a deepfake detection system \cite{khan2024generalization, ameer2025augmentation}. In line with this, multiple data augmentation techniques (such as JPEG compression, random resizing, cropping, rotations, horizontal flipping, colour jitter, Gaussian noise, and Gaussian blur) are applied during training. Overall, the primary goal of augmentation is to increase the data variety by applying different techniques to simulate the variation and diversity of real-world images. 

\section{Experimental Details}
\label{section:Details-of-Experiments}
The following set of experiments is conducted to assess the performance and generalizability of the proposed framework.
\begin{enumerate}
    \item intra-dataset performance evaluation
    \item cross-model evaluation (with and without augmentation)
    \item cross-paradigm evaluation (with and without augmentation)
    \item multi-source evaluation (with and without augmentation)
    \item real-world evaluation
\end{enumerate}

The framework is implemented using the PyTorch DL framework and trained on a system equipped with NVIDIA GeForce RTX 3060 (12 GB VRAM). The software environment included Python 3.9.25 and CUDA 13.1 for GPU acceleration. All source code is publicly available on \href{https://github.com/amnaamjid/multi-domain-deepfake-detection.git}{GitHub} for the benefit of the research community.

During training, the model is optimized using the AdamW optimizer with an initial learning rate of 0.001, which is then reduced to 0.0001. The batch size is set to 32, and training is performed over 30 epochs with a weight decay factor of 0.0001 applied using the AdamW optimizer to prevent overfitting and improve generalization. A ReduceLROnPlateau scheduler is used to monitor the learning rate, and it is reduced by a factor of 0.5 when the validation loss plateaus for 3 consecutive epochs. Cross-Entropy Loss with label smoothing (0.05) and a 0.25 dropout rate are used to prevent overfitting.

The metrics used to assess performance include accuracy, precision, recall, F1-score, ROC-AUC (Area Under the Receiver Operating Characteristic Curve), and Equal Error Rate (EER).

\begin{table*}[!t]
\caption{Intra-dataset evaluation results of SGFF-Net, where training and testing are performed on the same dataset. Results are reported as percentages (\%) except inference time, which is reported in milliseconds (ms). Higher values indicate better performance for Accuracy, Precision, Recall, F1-Score, and ROC-AUC, whereas lower values are preferred for EER and inference time.}
\label{tab:intra_dataset_evaluation}
\centering
\small
\setlength{\tabcolsep}{5pt}

\begin{tabular*}{\textwidth}{@{\extracolsep{\fill}}lccccccc@{}}
\toprule
\textbf{Dataset} & \textbf{Acc.$\uparrow$} & \textbf{Prec.$\uparrow$} & \textbf{Rec.$\uparrow$} & \textbf{F1$\uparrow$} & \textbf{ROC-AUC$\uparrow$} & \textbf{EER$\downarrow$} & \textbf{Time (ms)$\downarrow$} \\
\midrule

DFF\_C   & 98.47 & 98.52 & 98.40 & 98.47 & 99.85 & 1.54 & 10.30 \\
DFF\_A   & \textbf{98.95} & \textbf{99.59} & 98.30 & \textbf{98.94} & \textbf{99.94} & \textbf{0.92} & 10.41 \\
DFF\_B   & 97.98 & 97.46 & 98.53 & 97.99 & 99.76 & 1.93 & \textbf{10.29} \\
DFFD\_A  & 98.89 & 98.99 & \textbf{98.80} & 98.89 & \textbf{99.94} & 1.04 & 10.41 \\
MIX\_ALL & 97.83 & 93.66 & 98.10 & 95.84 & 99.32 & 3.53 & 10.50 \\

\bottomrule
\end{tabular*}
\end{table*}


\section{Results}
\label{section:Results}
The proposed Spatial-Gradient-Frequency Fusion Network (SGFF-Net) framework is evaluated under progressively more challenging settings to assess both detection performance and generalization capability. First, intra-dataset experiments measure performance when training and testing are performed on the same data distribution. Next, cross-model and cross-paradigm evaluations assess robustness to unseen generators and generation paradigms. Multi-source experiments investigate the impact of training data diversity, while real-world evaluation measures generalization to unseen samples collected outside the benchmark datasets. Together, these experiments provide a comprehensive analysis of the effectiveness and robustness of SGFF-Net.

\subsection{Intra-dataset Performance Evaluation}
\label{subsection:Intra-dataset-Evaluation}
Table~\ref{tab:intra_dataset_evaluation} presents the results for training and testing on the same dataset. One key observation from this table is that SGFF-Net, which combines multi-feature fusion with dual residual feature learning, is highly effective for both diffusion-based and GAN-based datasets, with accuracy consistently exceeding 97.80\%.

Among all diffusion-based datasets, DFF\_A achieves the highest accuracy of 98.95\%. This is followed by DFF\_C (98.47\%) and DFF\_B (97.98\%), indicating that the SGFF-Net has effectively captured the underlying discriminative artifacts present in diffusion-based generators. As for the GAN-based datasets, SGFF-Net is also effective, achieving an accuracy of 98.89\% on DFFD\_A and a robust F1-score of 98.89\%.

In the case of the MIX\_ALL dataset, which combines GAN-based and diffusion-based generators, the accuracy is slightly lower at 97.83\%. There is a more noticeable drop in precision (93.66\%), and an increase in EER (3.53\%). This can be attributed to combining diverse generation techniques, which makes the classification task more challenging by adding complexity and variability to the data.

\begin{table*}[!t]
\caption{Cross-model evaluation of SGFF-Net. Models are trained on one generative model and evaluated on unseen models within the same generation paradigm. Specifically, DFF\_C is evaluated on the average performance across DiffFace subsets (A--E), while DFFD\_A is evaluated on DFFD\_B. Results are reported with and without augmentation. All metrics are reported as percentages (\%) except inference time, which is reported in milliseconds (ms). Higher values indicate better performance for Accuracy, Precision, Recall, F1-Score, and ROC-AUC, whereas lower values are preferred for EER and inference time.}
\label{tab:cross_model_evaluation}
\centering
\small
\setlength{\tabcolsep}{5pt}

\begin{tabular*}{\textwidth}{@{\extracolsep{\fill}}llccccccc@{}}
\toprule
\textbf{Train Set} & \textbf{Test Set} & \textbf{Acc.$\uparrow$} & \textbf{Prec.$\uparrow$} & \textbf{Rec.$\uparrow$} & \textbf{F1$\uparrow$} & \textbf{ROC-AUC$\uparrow$} & \textbf{EER$\downarrow$} & \textbf{Time (ms)$\downarrow$} \\
\midrule

\multicolumn{9}{c}{\textbf{Without Augmentation}} \\
\midrule

DFF\_C & DiffFace (Mean A--E) & 70.46 & 69.68 & 86.57 & 71.38 & 68.95 & 35.92 & 10.35 \\
DFFD\_A & DFFD\_B & 91.31 & 83.10 & 99.14 & 89.14 & 96.66 & 8.02 & 10.37 \\

\midrule

\multicolumn{9}{c}{\textbf{With Augmentation}} \\
\midrule

DFF\_C & DiffFace (Mean A--E) & \textbf{79.80} & \textbf{79.15} & \textbf{90.80} & \textbf{80.50} & \textbf{81.67} & \textbf{25.67} & 10.29 \\
DFFD\_A & DFFD\_B & \textbf{93.41} & \textbf{89.10} & \textbf{99.23} & \textbf{92.34} & \textbf{98.78} & \textbf{7.02} & \textbf{10.24} \\

\bottomrule
\end{tabular*}
\end{table*}

\subsection{Cross-Model Evaluation}
\label{subsection:Cross-Model-Evaluation}
The results in Table~\ref{tab:cross_model_evaluation} indicate that when SGFF-Net is trained on diffusion-based datasets, it exhibits a substantial drop in accuracy compared to the high intra-dataset accuracy of 98.95\%. Likewise, precision and ROC-AUC also decline significantly, indicating reduced discriminative capability. It should be noted that different diffusion generative models, such as unconditional ADM, DDPM, LDM, and PNDM, introduce distinct artifact distributions different from those introduced by conditional models such as SD v1.5 text-to-image and SD Inpainting. These differences limit the transferability of features learned from a single source.

In contrast, the performance of GAN-based cross-model generalization is better with SGFF-Net achieving an accuracy of 91.31\% and an F1-score of 89.14\% when trained on DFFD\_A and tested on the DFFD\_B dataset. It also obtains the highest ROC-AUC (96.66\%) and the lowest EER of 8.02\%, indicating the effectiveness of SGFF-Net in capturing transferable features across GAN-based generators.

Additionally, results in Table~\ref{tab:cross_model_evaluation} indicate a noticeable difference in performance with and without augmentation. In the case of diffusion-based cross-model evaluation (DFF\_C $\rightarrow$ DiffFace), augmentation boosts accuracy from 70.46\% to 79.80\%, precision from 69.68\% to 79.15\%, recall from 86.57\% to 90.80\%, and F1-score from 71.38\% to 80.50\%. Similarly, augmentation also improves the results for GAN-based cross-model evaluation (DFFD\_A $\rightarrow$ DFFD\_B). For example, accuracy improves from 91.31\% to 93.41\%, and precision from 83.10\% to 89.10\%.

\begin{table*}[!t]
\caption{Cross-paradigm evaluation of SGFF-Net under bidirectional training and testing settings. The model is trained on diffusion-based datasets and evaluated on unseen GAN-based datasets, and vice versa. All metrics are reported as percentages (\%) except inference time, which is reported in milliseconds (ms). Higher values indicate better performance for Accuracy, Precision, Recall, F1-Score, and ROC-AUC, whereas lower values are preferred for EER and inference time.}
\label{tab:cross_paradigm_evaluation}
\centering
\small
\setlength{\tabcolsep}{5pt}

\begin{tabular*}{\textwidth}{@{\extracolsep{\fill}}llccccccc@{}}
\toprule
\textbf{Train Set} & \textbf{Test Set} & \textbf{Acc.$\uparrow$} & \textbf{Prec.$\uparrow$} & \textbf{Rec.$\uparrow$} & \textbf{F1$\uparrow$} & \textbf{ROC-AUC$\uparrow$} & \textbf{EER$\downarrow$} & \textbf{Time (ms)$\downarrow$} \\\midrule
\multicolumn{9}{c}{\textbf{Diffusion $\rightarrow$ GAN (Without Augmentation)}} \\ \midrule
\multirow{2}{*}{DFF\_C}
& DFFD\_A & 69.94 & 65.40 & 80.77 & 66.58 & 65.51 & 39.03 & 10.39 \\
& DFFD\_B & 65.25 & 64.62 & 79.76 & 66.73 & 67.80 & 43.71 & \textbf{10.30} \\ \midrule

\multicolumn{9}{c}{\textbf{Diffusion $\rightarrow$ GAN (With Augmentation)}} \\ \midrule

\multirow{2}{*}{DFF\_C}
& DFFD\_A & \textbf{78.12} & \textbf{74.25} & \textbf{86.10} & \textbf{78.18} & \textbf{84.27} & \textbf{24.15} & 10.42 \\
& DFFD\_B & 75.48 & 73.14 & 84.56 & 76.81 & 82.63 & 26.78 & \textbf{10.37} \\ \midrule

\multicolumn{9}{c}{\textbf{GAN $\rightarrow$ Diffusion (Without Augmentation)}} \\ \midrule

\multirow{3}{*}{DFFD\_A}
& DFF\_A & 69.62 & \textbf{92.17} & 57.03 & 54.24 & 72.34 & 31.80 & 10.41 \\
& DFF\_B & 63.85 & 61.74 & 50.22 & 60.46 & 58.96 & 42.80 & 10.42 \\
& DFF\_C & 66.73 & 74.24 & 60.62 & 62.28 & 65.64 & 36.61 & \textbf{10.39} \\ \midrule

\multicolumn{9}{c}{\textbf{GAN $\rightarrow$ Diffusion (With Augmentation)}} \\ \midrule

\multirow{3}{*}{DFFD\_A}
& DFF\_A & \textbf{78.15} & 85.42 & \textbf{74.28} & \textbf{77.69} & \textbf{84.21} & \textbf{22.84} & 10.43 \\
& DFF\_B & 75.46 & 79.11 & 71.82 & 74.80 & 81.76 & 24.93 & \textbf{10.41} \\
& DFF\_C & 76.98 & \textbf{81.27} & 73.55 & 75.86 & 82.68 & 23.61 & 10.42 \\
\bottomrule
\end{tabular*}
\end{table*}

\subsection{Cross-Paradigm Evaluation}
\label{subsection:Cross-Paradigm-Evalution}
To comprehensively analyse the ability of SGFF-Net to generalize across fundamentally different generative paradigms, two complementary experiments are conducted:
\begin{enumerate}
    \item training on diffusion-based datasets and testing on GAN-based datasets, and
    \item training on GAN-based datasets and testing on diffusion-based datasets.
\end{enumerate}

In the first experiment, the SGFF-Net is trained on the diffusion-based dataset DFF\_C and tested on different GAN-based datasets, such as DFFD\_A and DFFD\_B. Table~\ref{tab:cross_paradigm_evaluation} shows that the framework achieves accuracies of 69.94\% and 65.25\% on DFFD\_A and DFFD\_B, respectively. This is a significant drop in performance compared to intra-dataset and cross-model evaluation. The results also show an increase in false positives, as recall remains relatively high (80.77\%), while precision drops to 65.40\% when tested on DFFD\_A. With augmentation, the accuracies increase to 78.12\% and 75.48\% for DFFD\_A and DFFD\_B, respectively. Similarly, ROC-AUC values improve and EER values decrease, highlighting the benefit of using augmentation.

In the second experiment, SGFF-Net is trained on the GAN-based DFFD\_A dataset and tested on multiple diffusion-based datasets, including DFF\_A, DFF\_B, and DFF\_C. The results are presented in Table~\ref{tab:cross_paradigm_evaluation}, and they reveal a decline in generalization performance, with the SGFF-Net achieving accuracies of 69.62\%, 63.85\%, and 66.73\% on DFF\_A, DFF\_B, and DFF\_C, respectively. In contrast, after augmentation, the accuracies increase to 78.15\%, 75.46\%, and 76.98\%, respectively.

\begin{table*}[!t]
\caption{Generalization performance of SGFF-Net under mixed-source training and testing settings. Models trained on a single source are evaluated on the mixed-source test set (MIX\_ALL), while models trained on MIX\_ALL are evaluated on individual source-specific test sets. MIX\_ALL combines multiple diffusion- and GAN-generated datasets. Results demonstrate that mixed-source training achieves near-perfect generalization across individual sources, whereas augmentation substantially improves the robustness of single-source models on mixed-source data. All metrics are reported as percentages (\%) except inference time, which is reported in milliseconds (ms).}
\label{tab:multi_source_evaluation}
\centering
\small
\setlength{\tabcolsep}{5pt}

\begin{tabular*}{\textwidth}{@{\extracolsep{\fill}}llccccccc@{}}
\toprule
\textbf{Train Set} & \textbf{Test Set} & \textbf{Acc.$\uparrow$} & \textbf{Prec.$\uparrow$} & \textbf{Rec.$\uparrow$} & \textbf{F1$\uparrow$} & \textbf{ROC-AUC$\uparrow$} & \textbf{EER$\downarrow$} & \textbf{Time (ms)$\downarrow$} \\
\midrule

\multicolumn{9}{c}{\textbf{Single-Source $\rightarrow$ Mixed-Source (Without Augmentation)}} \\
\midrule

DFF\_C & MIX\_ALL & 78.48 & 75.95 & 82.85 & 79.79 & 85.31 & 23.19 & 10.51 \\
DFFD\_A & MIX\_ALL & 70.33 & 84.50 & 79.79 & 69.66 & 79.17 & 27.48 & \textbf{10.39} \\

\midrule

\multicolumn{9}{c}{\textbf{Single-Source $\rightarrow$ Mixed-Source (With Augmentation)}} \\
\midrule

DFF\_C & MIX\_ALL & \textbf{84.27} & 81.64 & \textbf{87.32} & \textbf{84.15} & \textbf{90.76} & \textbf{16.72} & 10.54 \\
DFFD\_A & MIX\_ALL & 78.19 & \textbf{86.28} & 83.51 & 80.44 & 87.45 & 19.86 & \textbf{10.42} \\

\midrule

\multicolumn{9}{c}{\textbf{Mixed-Source $\rightarrow$ Single-Source}} \\
\midrule

\multirow{4}{*}{MIX\_ALL}
& DFF\_C & \textbf{99.69} & 99.06 & 99.68 & \textbf{99.37} & \textbf{99.96} & \textbf{0.52} & \textbf{10.21} \\
& DiffFace\_A & 99.14 & \textbf{99.51} & 98.53 & 99.02 & 99.90 & 0.99 & 10.32 \\
& DFFD\_A & 99.51 & 97.44 & \textbf{99.83} & 98.62 & \textbf{99.96} & 0.55 & 10.48 \\
& DFFD\_B & 99.43 & 92.87 & \textbf{99.88} & 96.24 & 99.87 & 0.96 & 10.29 \\

\bottomrule
\end{tabular*}
\end{table*}

\subsection{Multi-source Generalization Evaluation}
\label{subsection:Multi-source-Genralisation-Evaluation}
To assess the effectiveness of multi-source training for improving generalization capability, two experimental settings are considered:
\begin{enumerate}
    \item training on a single dataset and testing on a mixed dataset, and
    \item training on a mixed dataset and evaluating using multiple single datasets.
\end{enumerate}

The results from the first experiment are shown in Table~\ref{tab:multi_source_evaluation}. It can be seen that training on a diffusion-based dataset and testing on a mixed dataset yields better generalization compared to when training is done using GAN-based datasets. The difference in accuracy is about 8\%, and when augmentation is used, the accuracy can be as high as 84.27\%.

The results from the second experiment, in which SGFF-Net is trained on the MIX\_ALL dataset and tested across individual datasets (including DFF\_C, DiffFace, DFFD\_A, and DFFD\_B), are shown in Table~\ref{tab:multi_source_evaluation}. There is a substantial improvement in performance, with accuracy exceeding 99\% on all datasets with multi-source training distributions.

Overall, Table~\ref{tab:multi_source_evaluation} leads to a few insights. Firstly, generalization to truly unseen distributions remains a challenging problem, but data diversity can help. More specifically, the results indicate that adding a small number of samples into the training set from the generative approaches and models used to generate the samples in the test set is an effective technique. It substantially reduces domain gaps that occur during single-source training; however, the remaining performance degradation on real-world data suggests that such gaps are not eliminated.

\begin{table*}[!t]
\caption{Real-world evaluation results measuring the out-of-distribution generalization capability of SGFF-Net on an unseen real-world deepfake benchmark containing images generated by diverse generation models and post-processing pipelines that are not present in the training data. Models are trained on diffusion-based, GAN-based, or mixed-source datasets and evaluated exclusively on the real-world test set. Results are reported as percentages (\%) except inference time, which is reported in milliseconds (ms). Higher values indicate better performance for Accuracy, Precision, Recall, F1-Score, and ROC-AUC, whereas lower values are preferred for EER and inference time.}
\label{tab:real_world_evaluation}
\centering
\small
\setlength{\tabcolsep}{5pt}

\begin{tabular*}{\textwidth}{@{\extracolsep{\fill}}llccccccc@{}}
\toprule
\textbf{Train Set} &
\textbf{Aug.} &
\textbf{Acc.$\uparrow$} &
\textbf{Prec.$\uparrow$} &
\textbf{Rec.$\uparrow$} &
\textbf{F1$\uparrow$} &
\textbf{ROC-AUC$\uparrow$} &
\textbf{EER$\downarrow$} &
\textbf{Time (ms)$\downarrow$} \\
\midrule

DFFD\_A  & No  & 45.00 & 43.90 & 36.00 & 39.56 & 42.96 & 45.05 & 14.55 \\
DFF\_C   & No  & 49.00 & 47.47 & 75.00 & 58.14 & 48.47 & 51.00 & 13.06 \\
MIX\_ALL & No  & 61.50 & 60.59 & 62.00 & 63.29 & 61.92 & 41.00 & \textbf{12.64} \\
MIX\_ALL & Yes & \textbf{75.80} & \textbf{70.80} & \textbf{85.00} & \textbf{73.08} & \textbf{77.86} & \textbf{31.00} & 13.44 \\

\bottomrule
\end{tabular*}
\end{table*}

\subsection{Real-World (Unseen) Data Evaluation}
\label{subsection:Generalization-to-Real-World-Data}
The advantage of using multi-domain feature learning and multi-source training is clearly evident from the results in Table~\ref{tab:real_world_evaluation}, in which the Real\_World dataset is used for testing. Specifically, when SGFF-Net utilises the MIX\_ALL dataset with augmentation for training, the highest accuracy of 75.80\% is achieved. In contrast, when a single-source dataset is used for training, the accuracy drops to a low of 49.00\%.

Additionally, the reduction in EER from 51.00\% (single-source training dataset) to 31.00\% (multi-source training dataset) suggests that augmentation techniques, along with multi-source training data, introduce additional variability, which helps SGFF-Net learn more generalized features. This observation indicates that generalization to real-world data requires both diversity and variability in training conditions, making it data and source-dependent.

\begin{figure*}[!t]
\centering
\includegraphics[width=0.6\linewidth]{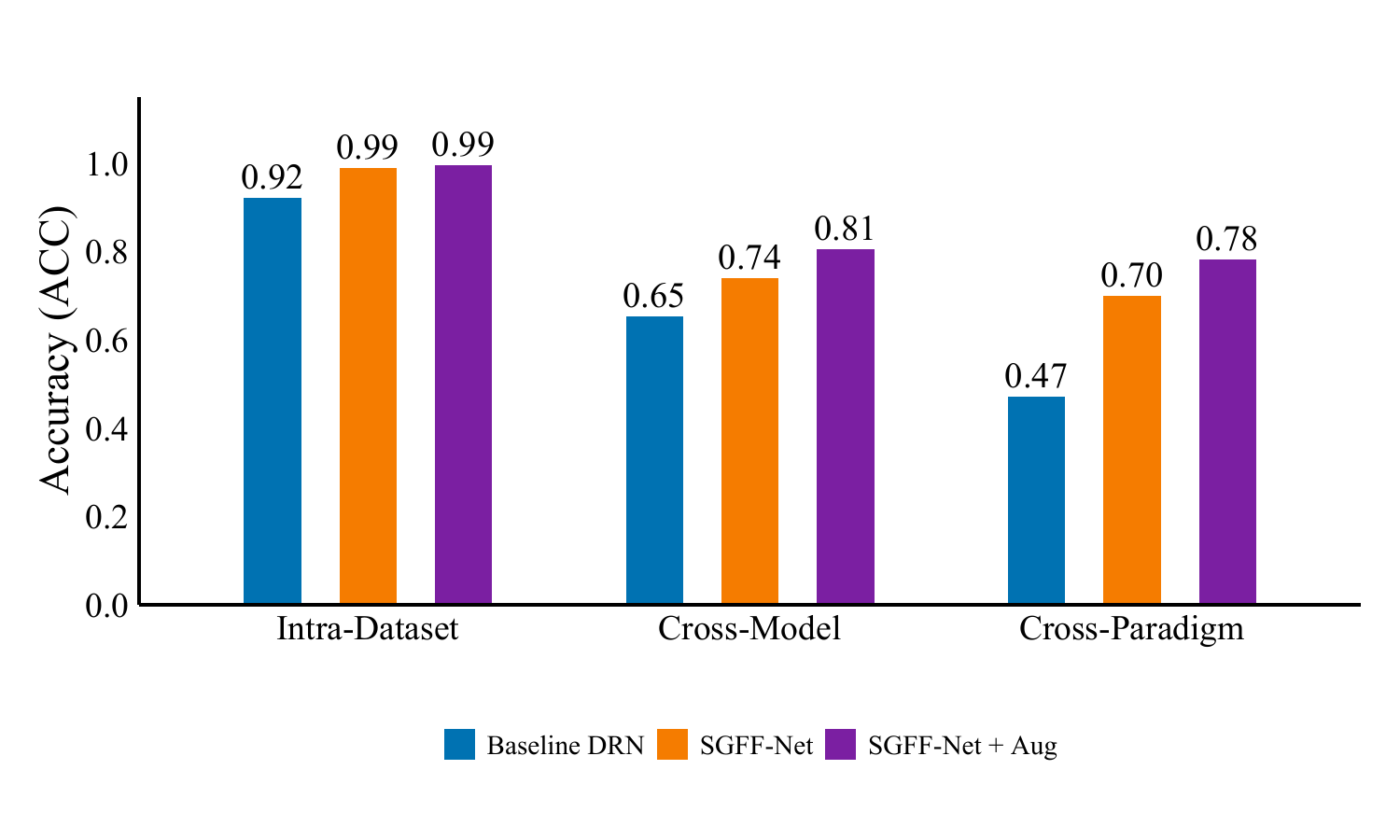}
\caption{Accuracy comparison of the baseline DRN, SGFF-Net, and the SGFF-Net with augmentation across intra-dataset, cross-model, and cross-paradigm evaluation settings.}
\label{fig:accuracy_comparasion}
\end{figure*}

\begin{figure*}[!t]
\centering
\includegraphics[width=0.6\linewidth]{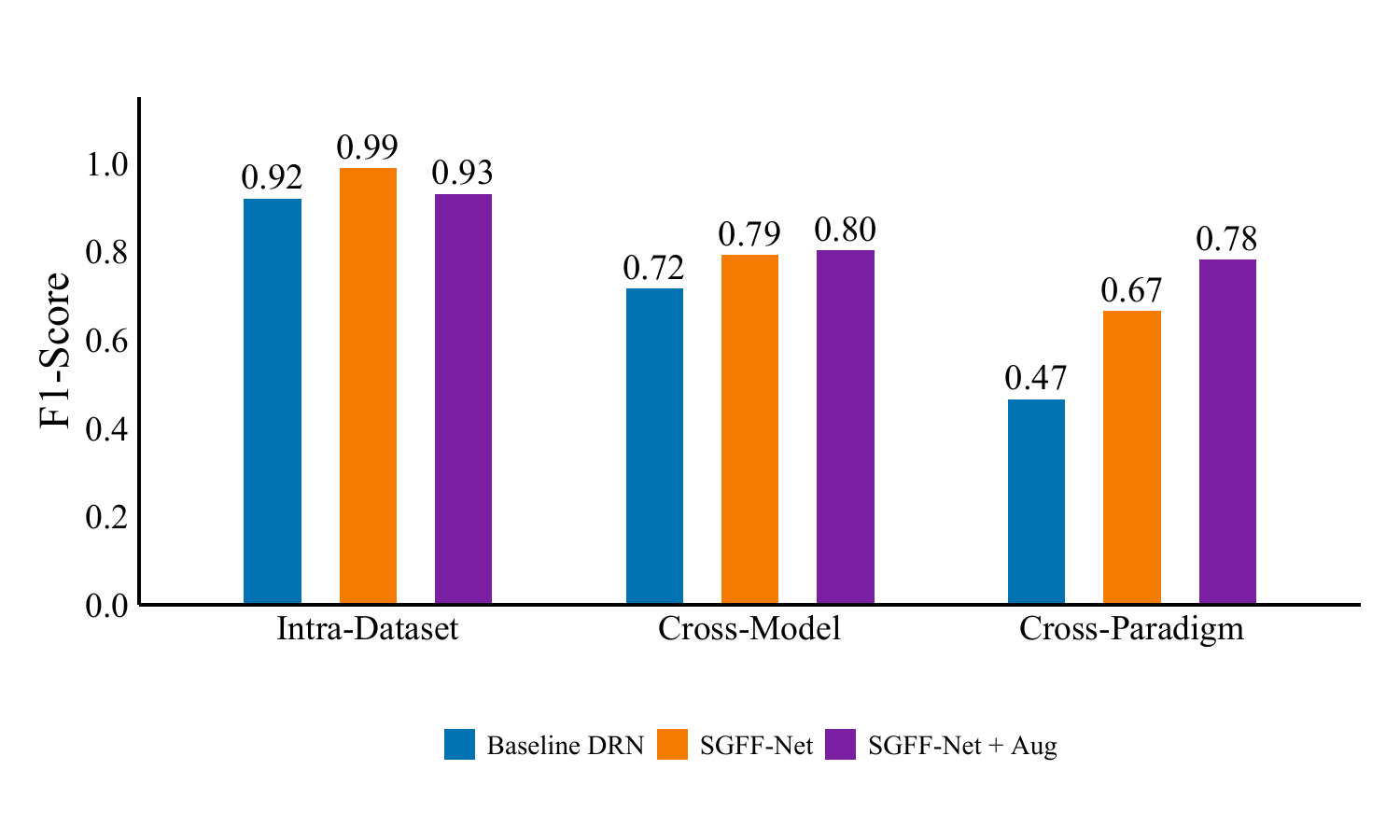}
\caption{F1-score comparison of the baseline DRN, SGFF-Net, and the SGFF-Net with augmentation across intra-dataset, cross-model, and cross-paradigm evaluation settings.}
\label{fig:F1_Comparison}
\end{figure*}

\begin{figure*}[!t]
\centering
\includegraphics[width=0.6\linewidth]{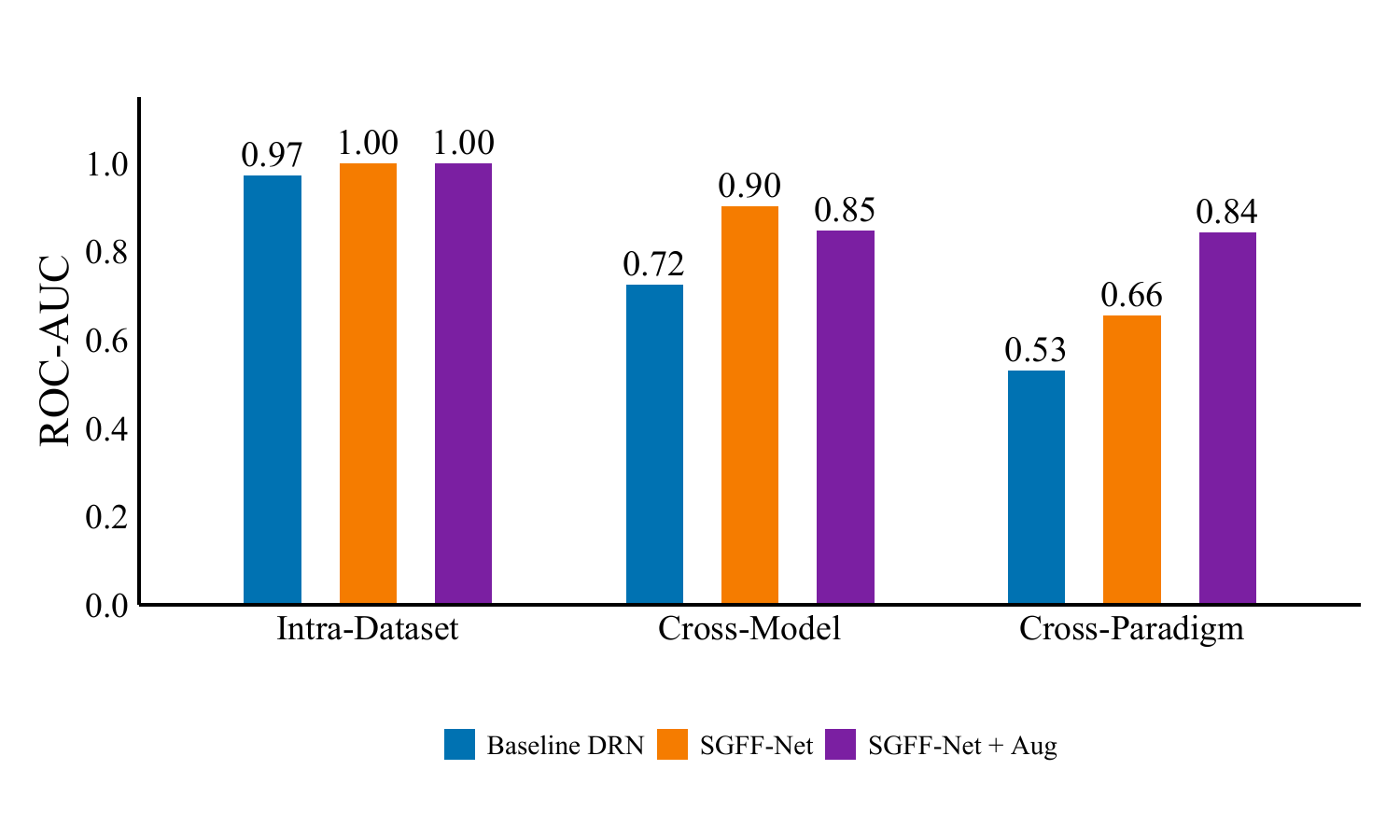}
\caption{ROC-AUC comparison of the baseline DRN, SGFF-Net, and the SGFF-Net with augmentation across intra-dataset, cross-model, and cross-paradigm evaluation settings.}
\label{fig:ROC_AUC_Comparison}
\end{figure*}


\section{Discussion}
\label{section:Discussion}
The experimental results demonstrate that SGFF-Net achieves strong performance across a wide range of evaluation settings, ranging from controlled intra-dataset testing to challenging cross-paradigm and real-world scenarios. While it substantially improves robustness compared to conventional DRN-based detectors, the results also reveal important insights regarding architectural design choices, the nature of domain shift across generative models, and the role of data diversity and augmentation in achieving generalization. This section synthesizes these findings and relates them to the design decisions of the proposed framework. Detailed ablation studies supporting the architectural choices are provided in Appendix~\ref{section:Ablation-Study}.

\subsection{Architectural Design Insights}
\label{subsection:Architectural-Design-Insights}
The architectural design of SGFF-Net is guided by a series of ablation studies. Detailed experimental results are reported in Appendix~\ref{section:Ablation-Study}, while the key findings are summarized here.

Although this study draws motivation from \cite{sharafudeen2025DRN}, the design of the proposed framework is developed in an incremental manner with design choices being made on the basis of empirical data. During the ablation study, different frequency representations (FFT, DCT, and DWT) and image representations (RGB and grayscale) are evaluated to determine which is best for detecting diffusion-generated deepfakes. The results show that DWT-based frequency representations consistently outperform FFT and DCT for both grayscale and RGB images (see Table~\ref{tab:frequency_ablation}). Compared with FFT, DWT improves accuracy by 5.17\% for grayscale images and 6.60\% for RGB images while simultaneously reducing EER. This is because the generation process of diffusion-based deepfakes introduces subtle and spatially varying artifacts during iterative denoising, which are better preserved by the multi-resolution analysis provided by DWT.

The same ablation study further shows that RGB inputs consistently outperform grayscale inputs, indicating that colour information contributes additional forensic cues beyond those captured by luminance information alone. This observation supports the decision to retain RGB representations within the final feature extraction module.

The ablation experiments also reveal that the effectiveness of the frequency branch depends not only on the use of DWT but also on the choice of wavelet family and boundary handling strategy. Among the evaluated configurations, Symlet wavelets combined with the Reflect boundary mode achieve the highest ROC-AUC, indicating that both the wavelet basis and border extension mechanism materially influence the discriminative quality of the extracted frequency-domain features.

In addition, the subband analysis reported in Appendix~\ref{section:Ablation-Study} demonstrates that combining all four DWT subbands (LL, LH, HL, and HH) consistently yields the strongest performance across same-dataset, cross-model, and cross-generator settings. This suggests that deepfake artifacts are distributed across multiple frequency scales rather than being concentrated within a single subband. Consequently, preserving information from all subbands allows the SGFF-Net to exploit both global structural information and localized high-frequency manipulation traces.

Finally, the feature fusion analysis demonstrates that the combination of RGB, gradient, and frequency branches consistently outperforms all individual branches and pairwise combinations. This finding indicates that spatial, gradient-based, and frequency-domain representations provide complementary information rather than redundant signals. Therefore, the final three-branch architecture adopted in SGFF-Net is empirically justified by the ablation results.

\subsection{Comparison with Baseline DRN}
\label{subsection:Comparison-with-Baseline-DRN}
For comparison, a baseline DRN similar to \cite{sharafudeen2025DRN} is also implemented and evaluated. Sharafudeen \& Chandra show that their DRN model (with grayscale and FFT) achieves near-perfect accuracy (99\%) and low EER (0.02) during intra-dataset testing \cite{sharafudeen2025DRN}. However, Fig.~\ref{fig:accuracy_comparasion} shows that the baseline DRN implemented in this paper (with DWT and RGB images) achieves an accuracy of 92.04\% and does not generalize well on unseen data. In fact, its accuracy drops sharply from 92.04\% to 47.19\%. 

A notable drop in accuracy is also observed during cross-model evaluation, where performance decreases from 92.04\% to 65.22\%. This behaviour indicates that the baseline DRN predominantly learns generator-specific artifacts rather than robust and transferable forensic cues. In contrast, the accuracy, F1-score, and ROC-AUC of SGFF-Net (shown in Figs.~\ref{fig:accuracy_comparasion}, \ref{fig:F1_Comparison}, and \ref{fig:ROC_AUC_Comparison}) are consistently higher across challenging evaluation settings. These results indicate that multi-domain feature fusion contributes to improved robustness against domain shift; however, the largest gains are obtained when feature diversity is complemented by data augmentation and multi-source training.

\subsection{Understanding Generalization Across Generators and Paradigms}
\label{subsection:Understanding-Generalization}
A clear hierarchy emerges across all evaluation settings. Performance is highest in intra-dataset evaluation, followed by cross-model evaluation, cross-paradigm evaluation, and finally real-world evaluation. This progressive degradation demonstrates that the difficulty of deepfake detection increases as the gap between training and testing distributions widens.

Although SGFF-Net significantly enhances deepfake detection performance, it does not eliminate the generalization gap. In fact, feature richness alone appears insufficient to achieve robust generalization under all conditions. The results from Tables~\ref{tab:cross_model_evaluation}, \ref{tab:cross_paradigm_evaluation}, \ref{tab:multi_source_evaluation}, and \ref{tab:real_world_evaluation} indicate that performance remains partially dependent on the distribution of the training data, and SGFF-Net continues to struggle when confronted with generators exhibiting fundamentally different generation characteristics.

The results further indicate that the domain gap between diffusion-based and GAN-based generators is substantially larger than the gap between generators within the same paradigm. This observation explains why performance degradation is more pronounced in cross-paradigm evaluation than in cross-model evaluation. Consequently, cross-paradigm evaluation remains one of the most challenging controlled settings despite the use of multi-domain features.

The real-world evaluation results provide further evidence that generalization remains an open problem. Even though SGFF-Net achieves substantially better performance than the baseline detector, the remaining gap between benchmark performance and real-world performance indicates that learned representations are not yet fully invariant to unseen data distributions. Therefore, improving generator-invariant feature learning remains an important research direction.

\subsection{Impact of Data Augmentation and Multi-Source Training}
\label{subsection:Impact-of-Augmentation}
The experimental results consistently demonstrate the importance of data diversity and training variability for improving generalization. Across cross-model, cross-paradigm, mixed-source, and real-world evaluations, the use of augmentation consistently improves accuracy, F1-score, ROC-AUC, and EER.

Augmentation helps by introducing variability such as noise, compression artifacts, illumination changes, and geometric transformations during training. This forces SGFF-Net to learn more robust and invariant feature representations rather than memorizing dataset-specific patterns. The effect is particularly evident in cross-model and cross-paradigm evaluations, where accuracy improvements of approximately 8--10\% are observed across multiple settings.

The results also demonstrate the importance of training data diversity. When the framework is trained on the MIX\_ALL dataset, which combines multiple diffusion-based and GAN-based generators, performance improves substantially across nearly all evaluation settings. The addition of even a limited number of samples from previously unseen generators reduces domain gaps and improves the transferability of learned representations. This observation highlights the importance of domain exposure in learning robust forensic features.

The strongest evidence of this effect appears in the real-world evaluation. Accuracy increases from 61.50\% to 75.80\% when augmentation is combined with mixed-source training, while EER decreases from 41.00\% to 31.00\%. These results suggest that feature diversity alone is insufficient; robust generalization requires a combination of feature diversity, data diversity, and training variability.

It should be noted here that although prior work 
(e.g., \cite{chen2024mcdm,ameer2025augmentation}) evaluated the impact of augmentation, it focused on conventional intra-dataset settings using DFDC \cite{dang2020detection} and Celeb-DF \cite{Li2020CelebDF} datasets. In contrast, this work evaluates augmentation under substantially more challenging conditions, including cross-model, cross-paradigm, multi-source, and real-world settings. Only Chen et al. reported an AUC improvement from 73.12\% to 78.02\% using diffusion-based augmentation \cite{chen2024mcdm}. Although direct comparison with their work is not possible due to differences in datasets, models, and augmentation strategies, the overall findings support the common conclusion that augmentation should not be viewed merely as a preprocessing step but rather as a critical component for improving robustness against unseen generative techniques.

\subsection{Practical Implications and Deployment Considerations}
\label{subsection:Practical-Implications}
The stable inference time observed across all evaluation settings demonstrates that the improvements achieved through feature fusion, augmentation, and multi-source training do not introduce substantial computational overhead. Across all experiments, inference remains approximately 10--14 ms per image, indicating that the framework is suitable for practical deployment in scenarios requiring both high detection performance and low latency.

Furthermore, the results suggest that real-world deployment should prioritize training data diversity in addition to architectural improvements. The substantial gains achieved through mixed-source training indicate that exposure to diverse generative processes is essential for building robust deepfake detectors capable of operating under continuously evolving threat landscapes.

\subsection{Limitations and Future Directions}
\label{subsection:Limitations}
Lastly, there are a few limitations in this study that should be stated. Firstly, the classification backbone (DRN) is inherently designed for single-domain spatial feature learning and might not fully exploit the complementary relationships between RGB, gradient, and frequency-domain representations. Secondly, the feature fusion strategy adopted in the proposed framework relies on simple concatenation, which does not explicitly model inter-feature interactions or attention-based weighting. Lastly, the frequency branch relies on fixed DWT decomposition without adaptive selection of informative sub-bands.

These limitations suggest several promising directions for future work. More advanced backbones based on attention mechanisms and transformers may better exploit cross-domain interactions among spatial, gradient, and frequency representations. Similarly, adaptive frequency analysis techniques, domain adaptation methods, and self-supervised learning approaches may further improve robustness to unseen generators. Finally, expanding the diversity of training data and designing explicitly domain-invariant representations remain important directions for achieving stronger real-world generalization.


\section{Conclusion}
\label{section:Conclusion}
This work investigates the detection of deepfake images generated by modern diffusion models and proposes SGFF-Net, a multi-domain feature fusion framework that integrates spatial (RGB), gradient, and frequency-domain representations through a multi-branch architecture and a DRN-based detector. The design of the framework is guided by extensive ablation studies, which identify DWT as the most effective frequency representation technique and demonstrate the complementary nature of spatial, gradient, and frequency-domain features.

Experimental evaluation across intra-dataset, cross-model, cross-paradigm, multi-source, and real-world settings demonstrates that SGFF-Net achieves strong detection performance while substantially improving robustness compared with a baseline DRN. The results show that multi-domain feature fusion improves the transferability of learned representations across different generators and generation paradigms. However, performance degradation under cross-paradigm and real-world evaluations confirms that domain shift remains a major challenge for deepfake detection.

The experiments further reveal that feature diversity alone is insufficient for achieving robust generalization. The largest gains are obtained when multi-domain feature learning is combined with data augmentation and multi-source training. In particular, mixed-source training substantially reduces generator-specific bias and improves robustness to unseen distributions, while augmentation consistently enhances performance across cross-model, cross-paradigm, and real-world evaluation settings. These findings suggest that effective generalization depends not only on feature design but also on training data diversity and variability.

Despite these improvements, real-world evaluation results indicate that current detectors remain sensitive to previously unseen generative processes and post-processing pipelines. Consequently, learning representations that remain invariant across evolving generators continues to be an open research problem.

Future work will explore more advanced architectures for modelling interactions among spatial, gradient, and frequency-domain representations. In particular, attention-based and transformer-based backbones such as Swin Transformer and ConvNeXt may provide more effective cross-domain feature integration. Additionally, domain adaptation, self-supervised learning, and contrastive representation learning offer promising directions for improving robustness to unseen distributions. Expanding training corpora to include a broader range of diffusion models and real-world manipulations, as well as developing explicitly domain-invariant forensic representations, constitute important directions for future investigation.

\bibliographystyle{IEEEtran}
\bibliography{references}

\appendices
\section{Ablation Study}
\label{section:Ablation-Study}

\setcounter{table}{0}
\renewcommand{\thetable}{\Alph{section}.\arabic{table}}
\renewcommand{\theHtable}{A\arabic{table}} 

\setcounter{figure}{0}
\renewcommand{\thefigure}{\Alph{section}.\arabic{figure}}
\renewcommand{\theHfigure}{A\arabic{figure}} 

This section describes the ablation study carried out to determine the best frequency representation method, DWT family, mode, and subband choice, to evaluate their effectiveness for the different feature extractor branches (RGB, gradient, and frequency).

\subsection{Selection of Frequency Representation Method}
\label{subsection:Selection-of-Frequency-Representation-Method}
Table ~\ref{tab:frequency_ablation} presents the results of using each frequency method on baseline DRN for RGB and greyscale images. The DRN model is trained and tested on the DFF\_A dataset. DWT achieved the best performance for both RGB and greyscale, with an accuracy of 92.04\% and an EER of 0.0798.

\subsubsection{Selection of DWT Subbands}
\label{subsubsection:Selection-of-DWT-Sub-bands}
\begin{figure*}[!htb]
\centering
\includegraphics[width=0.5\linewidth]{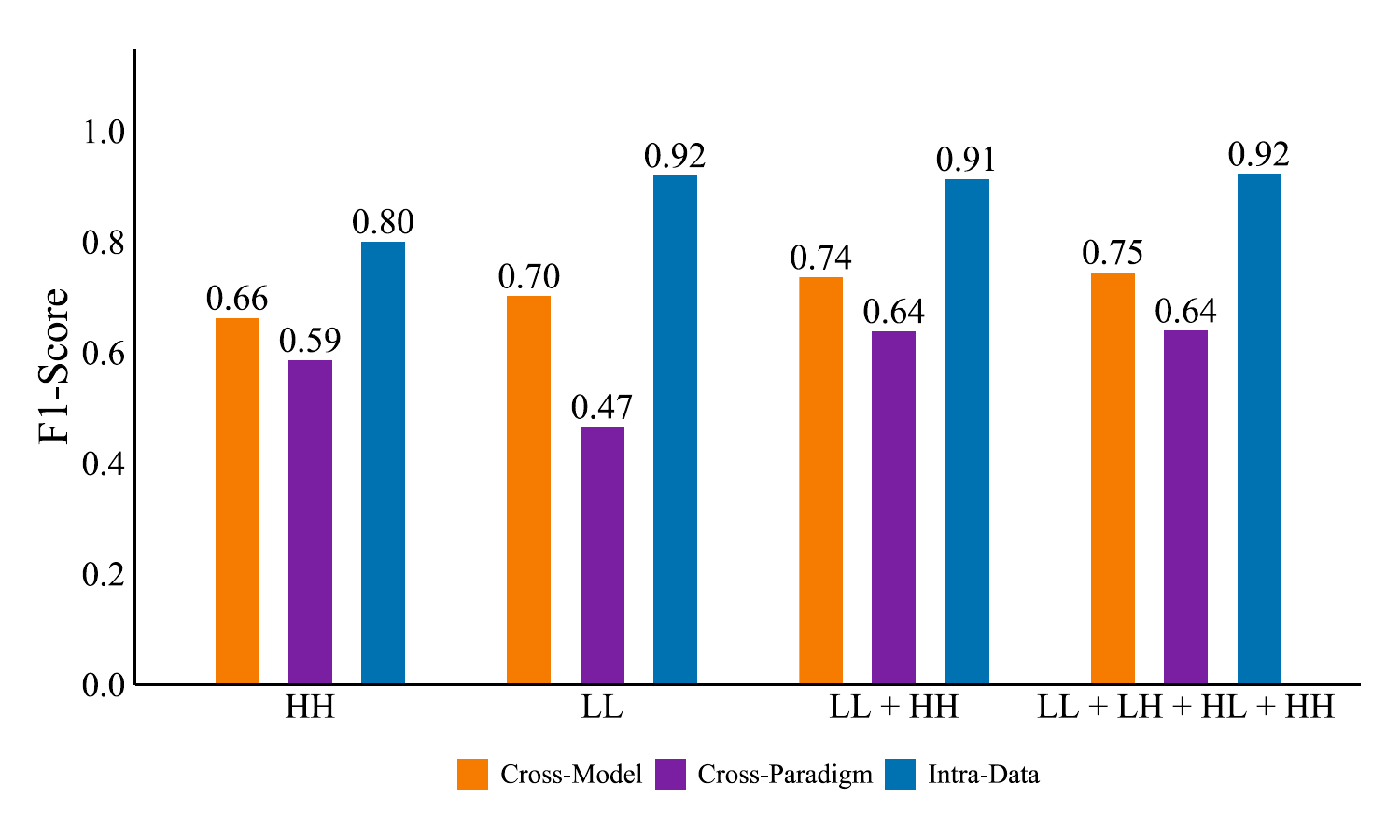}
\caption{F1-score comparison of different DWT subband configurations.}
\label{fig:f1_subband}
\end{figure*}

\begin{figure*}[!htb]
\centering
\includegraphics[width=0.5\linewidth]{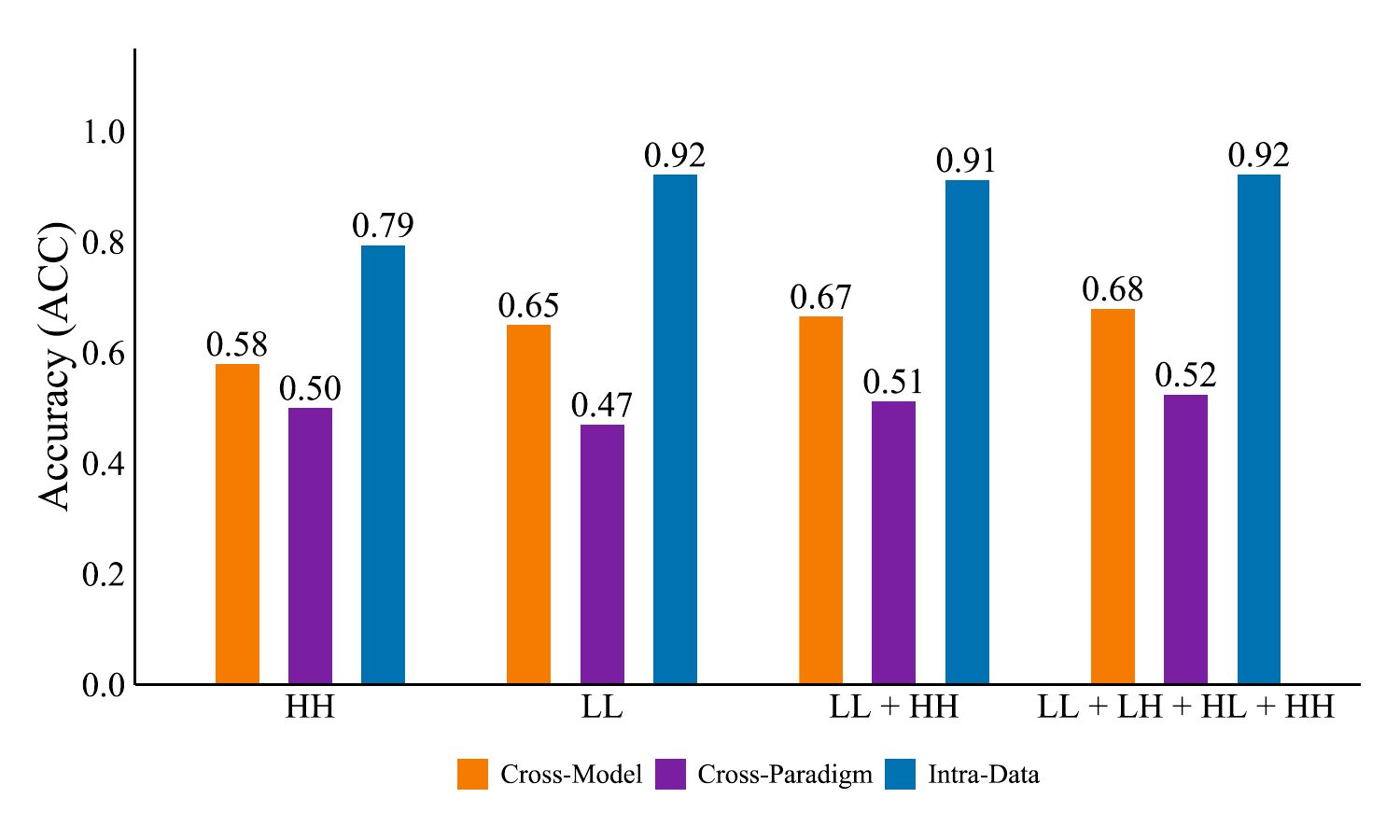}
\caption{Accuracy comparison of different DWT subband configurations.}
\label{fig:acc_subband}
\end{figure*}

2D DWT for images applies filtering in rows and columns, producing 4 subbands. Instead of testing all possible combinations, some combinations are selected for testing:  individual subbands (LL, HH), partial combinations (LL+HH), and full fusion of all subbands (LL+LH+HL+HH). The performance of each combination along with the baseline DRN is evaluated on the same-dataset, cross-dataset, and cross-generator settings. Figs. ~\ref{fig:f1_subband} and ~\ref{fig:acc_subband} show that the fusion of all subbands (LL+LH+HL+HH)) yielded the best results across all three settings.

\subsubsection{Selection of DWT Wavelet and Mode}
\label{subsubsection:Selection-of-DWT-Wavelet-and-Mode}
The Haar, Symlet, and Biorthogonal wavelet types are evaluated along with different signal extension modes (e.g., zero padding, symmetric, and anti-reflect). Table ~\ref{tab:wavelet_ablation} shows that the best AUC score is achieved when Symlet and anti-reflect are used.

\subsection{Selection of Feature Extractor Branches}
\label{subsection:Selection-of-Feature-Extractor-Branches}
\begin{figure*}[!htb]
\centering
\includegraphics[width=0.8\linewidth]{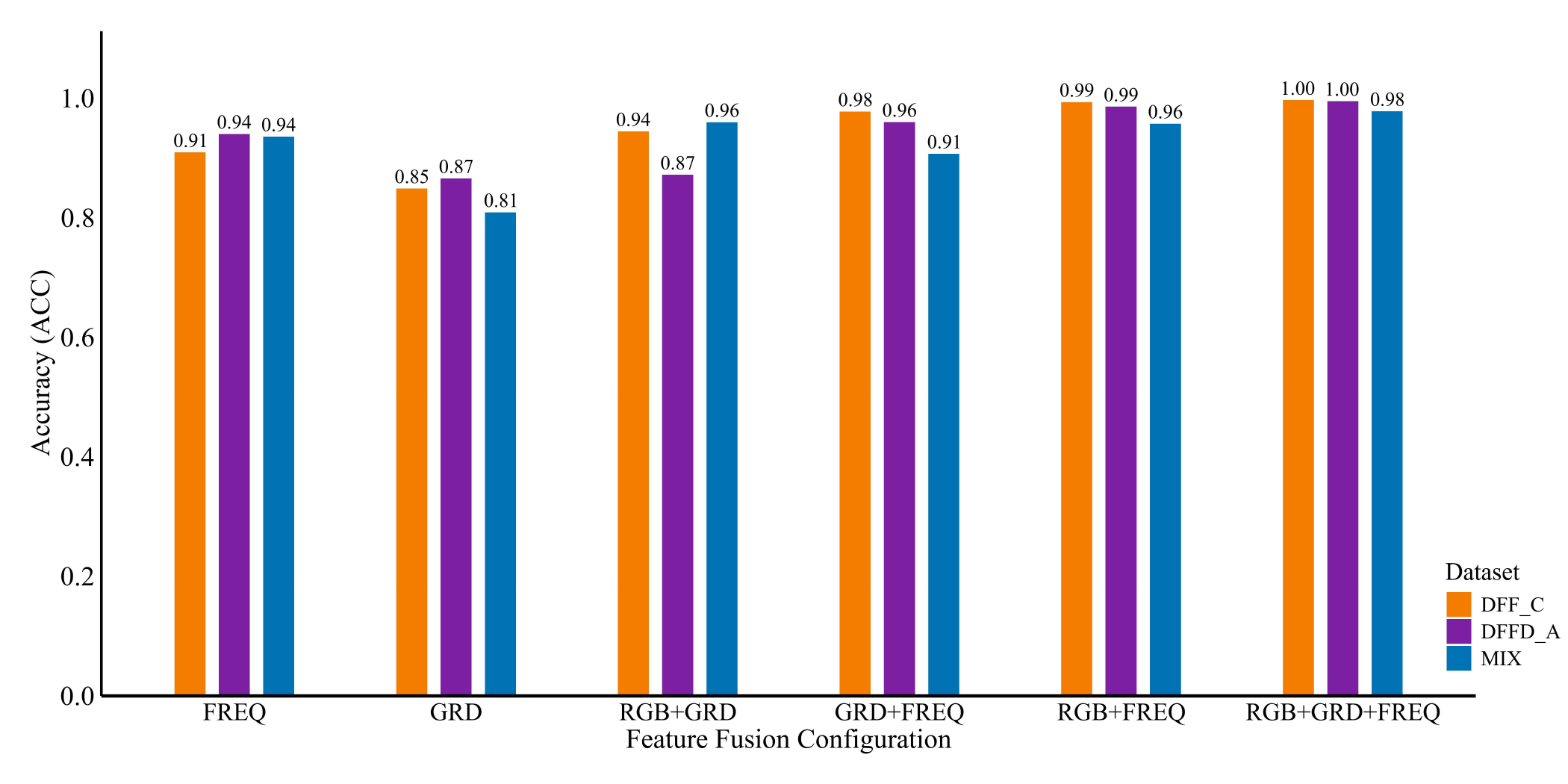}
\caption{Effect of feature fusion strategies across multiple test datasets.}
\label{fig:featureextractorfusion}
\end{figure*}

To design the feature extractor module, the contribution of the RGB branch, the gradient branch, and the frequency extractor branch are assessed. Also, several combinations of these branches are included in the assessment. Figure ~\ref{fig:featureextractorfusion} presents the accuracy obtained when the combinations are trained on mixed datasets and evaluated on different datasets individually. Clearly, RGB + GRD + FREQ has the highest accuracy and is included in the design shown in Fig. \ref{fig:architecture}.

\begin{table*}[!t]
\caption{Ablation study on frequency-domain representations using the DFF\_A dataset. The performance of Fast Fourier Transform (FFT), Discrete Cosine Transform (DCT), and Discrete Wavelet Transform (DWT) representations is evaluated using both grayscale and RGB inputs. Results are reported as percentages (\%) except inference time, which is reported in milliseconds (ms). Higher values indicate better performance for Accuracy, Precision, Recall, F1-Score, and ROC-AUC, whereas lower values are preferred for EER and inference time.}
\label{tab:frequency_ablation}
\centering
\small
\setlength{\tabcolsep}{5pt}

\begin{tabular*}{\textwidth}{@{\extracolsep{\fill}}llccccccc@{}}
\toprule
\textbf{Input Type} & \textbf{Method} & \textbf{Acc.$\uparrow$} & \textbf{Prec.$\uparrow$} & \textbf{Rec.$\uparrow$} & \textbf{F1$\uparrow$} & \textbf{ROC-AUC$\uparrow$} & \textbf{EER$\downarrow$} & \textbf{Time (ms)$\downarrow$} \\
\midrule

\multirow{3}{*}{Grayscale}
& FFT & 83.82 & 86.26 & 80.47 & 83.26 & 91.85 & 15.83 & 0.80 \\
& DCT & 79.39 & 76.06 & 85.79 & 80.63 & 87.77 & 20.18 & \textbf{0.58} \\
& DWT & \textbf{88.99} & \textbf{89.56} & \textbf{88.27} & \textbf{88.91} & \textbf{95.15} & \textbf{11.08} & 2.40 \\
\cmidrule(lr){2-9}
& \multicolumn{8}{l}{\textit{DWT vs. FFT: +5.17 Acc., +3.30 ROC-AUC, and -4.75 EER}} \\

\midrule

\multirow{3}{*}{RGB}
& FFT & 85.44 & 82.61 & 89.78 & 86.05 & 92.64 & 14.59 & \textbf{1.08} \\
& DCT & 82.28 & 81.52 & 83.47 & 82.49 & 89.91 & 17.72 & 1.74 \\
& DWT & \textbf{92.04} & \textbf{92.12} & \textbf{91.94} & \textbf{92.03} & \textbf{97.16} & \textbf{7.98} & 2.50 \\
\cmidrule(lr){2-9}
& \multicolumn{8}{l}{\textit{DWT vs. FFT: +6.60 Acc., +4.52 ROC-AUC, and -6.61 EER}} \\

\bottomrule
\end{tabular*}
\end{table*}

\begin{table*}[!t]
\caption{Ablation study of wavelet families and boundary extension modes. Results are reported using ROC-AUC (\%) on the DFF\_A dataset. Higher values indicate better performance.}
\label{tab:wavelet_ablation}
\centering
\small

\begin{tabular}{llc}
\toprule
\textbf{Wavelet Family} & \textbf{Boundary Mode} & \textbf{ROC-AUC (\%)$\uparrow$} \\
\midrule

Biorthogonal & Symmetric   & 91.42 \\
Biorthogonal & Reflect     & 91.97 \\
Biorthogonal & Antireflect & 92.97 \\

Symlet & Symmetric   & 93.87 \\
Symlet & Reflect     & \textbf{94.65} \\
Symlet & Antireflect & 93.21 \\

Haar & Symmetric   & 93.65 \\
Haar & Reflect     & 94.45 \\
Haar & Antireflect & 93.65 \\

\bottomrule
\end{tabular}
\end{table*}


\clearpage
\begin{IEEEbiography}[{\includegraphics[width=1in,height=1.25in,clip,keepaspectratio]{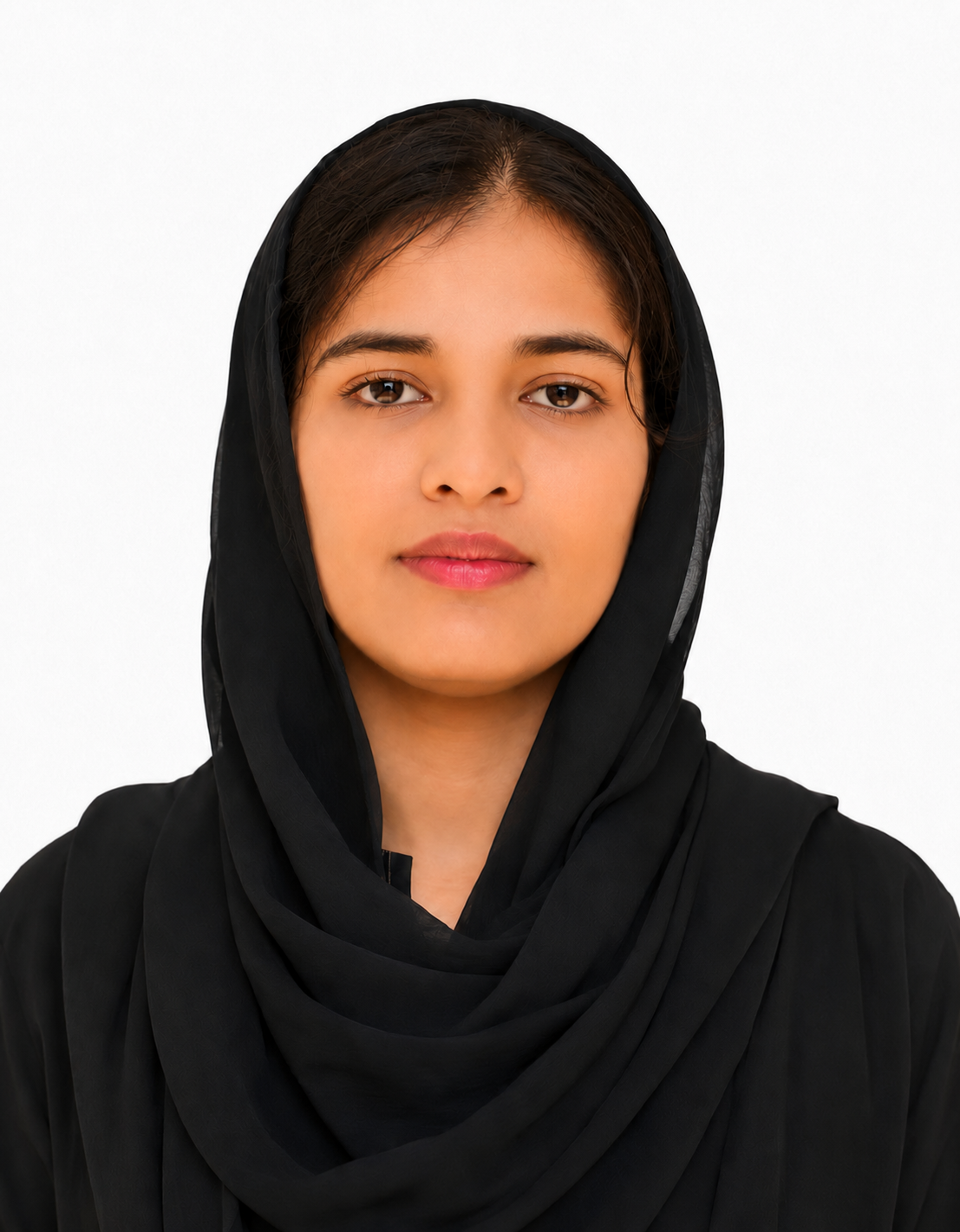}}]{Amna Amjid} received the B.S. degree in Software Engineering from the University of Haripur, Pakistan, in 2024, where she graduated as a Gold Medalist. She is currently pursuing the M.S. degree in Information Security at the School of Electrical Engineering and Computer Science (SEECS), National University of Sciences and Technology (NUST), Islamabad, Pakistan. Her interests include AI-driven security, deepfake detection, digital media forensics, malware analysis, and adversarial machine learning. Her current research focuses on robust detection of GAN- and diffusion-generated media using multi-domain image analysis.
\end{IEEEbiography}
\begin{IEEEbiography}[{\includegraphics[width=1in,height=1.25in,clip,keepaspectratio]{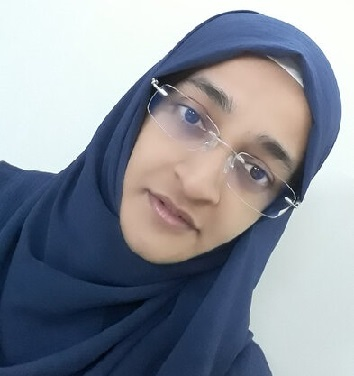}}]{Sana Qadir} completed her bachelor’s degree in Information Technology from the University of Southern Queensland, Australia, with Distinction. She then went on to complete an M.Sc. degree in Computer and Information Engineering and a Ph.D. degree in Engineering from International Islamic University Malaysia (IIUM), in 2010 and 2016, respectively. She is currently working as an Assistant Professor at the School of Electrical Engineering \& Computer Science (SEECS), NUST. She has more than 10 years of experience in industry and academia, and has served as an advisor for more than 20 research projects. Her research interests include network security, malware detection, digital forensics, security assessment, and applied cryptography. 
\end{IEEEbiography}

\begin{IEEEbiography}[{\includegraphics[width=1in,height=1.25in,clip,keepaspectratio]{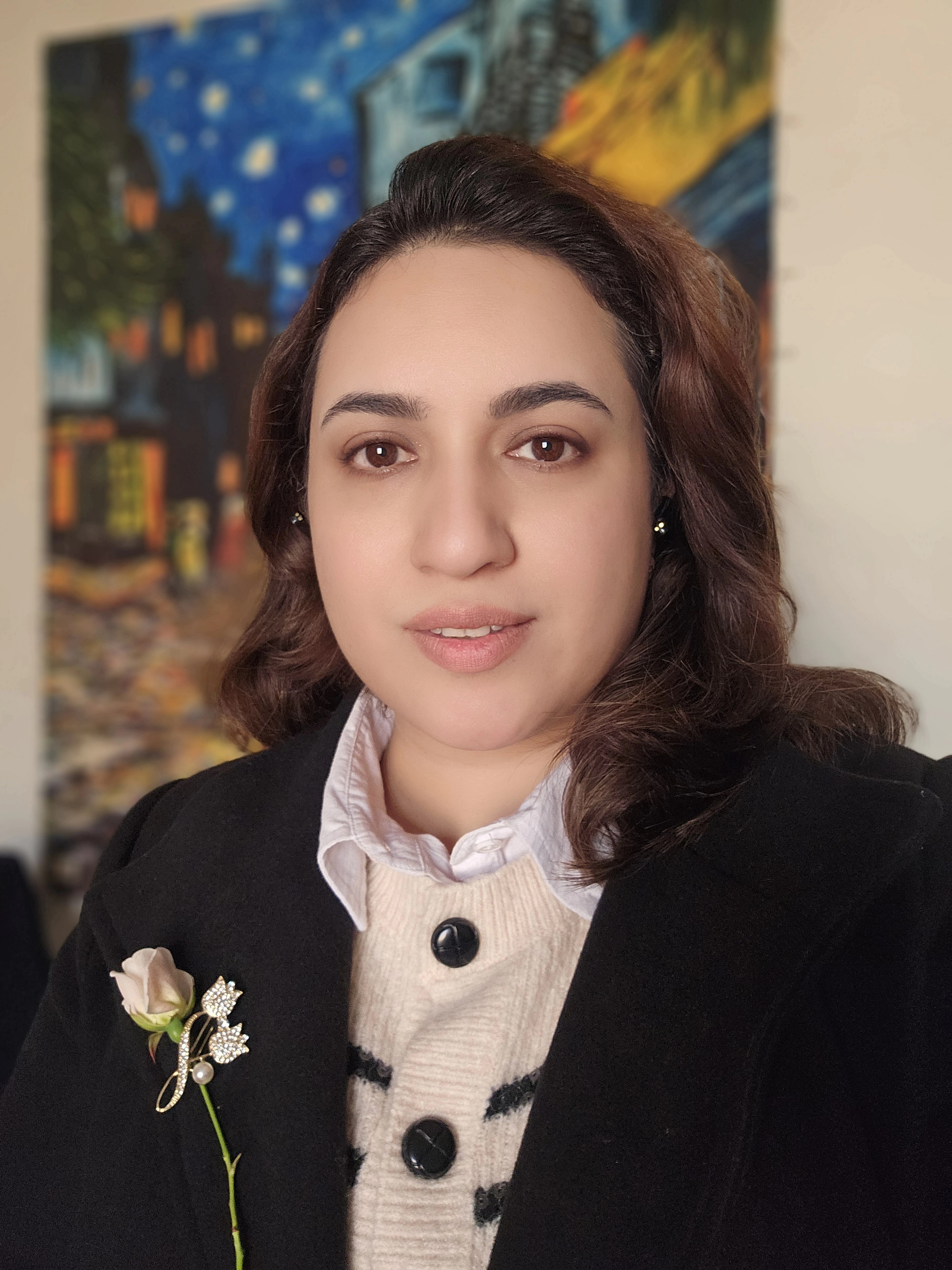}}]{Mehwish Fatima}is an AI researcher, educator, and industry practitioner with expertise in LLMs, generative AI, multimodal AI, and computational linguistics. She received the Ph.D. degree in Computational Linguistics from Heidelberg University, Germany, in 2024. She is currently an Assistant Professor with the Department of Artificial Intelligence and Data Science, School of Electrical Engineering and Computer Science (SEECS), National University of Sciences and Technology (NUST), Islamabad, Pakistan. She has research experience at the Heidelberg Institute for Theoretical Studies (HITS), Germany, and industry experience as a GenAI Solution and Content Architect with Educative Inc., USA. Her research interests include LLMs, multilingual and multimodal AI, explainable AI, and efficient machine learning. She has published in leading venues, including ACL and EMNLP, and serves as a reviewer for major conferences and journals in AI and NLP.
\end{IEEEbiography}
\begin{IEEEbiography}[{\includegraphics[width=1in,height=1.25in,clip,keepaspectratio]{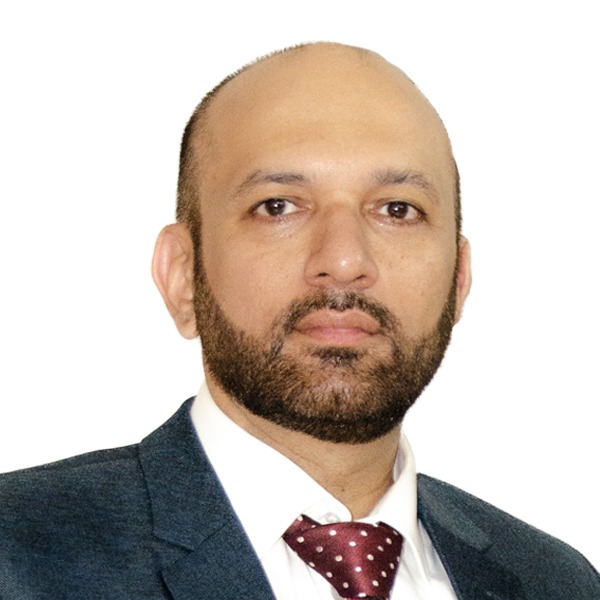}}]{Raja Khurram Shahzad} earned his PhD in Computer Science from Blekinge Institute of Technology and has more than 15 years of teaching and research experience in Sweden and Pakistan. His research spans cybersecurity, machine learning, trustworthy AI, and AI in education, with publications in peer-reviewed conferences and journals, including work on malware detection, application security, and secure systems. He also serves as a reviewer for various journals and as an associate editor, and he actively contributes to research supervision and academic development. He has led and contributed to funded projects and has taught and supervised at the undergraduate and graduate levels.
\end{IEEEbiography}
\EOD
\end{document}